\documentclass[manuscript,screen,review=false]{acmart}
\AtBeginDocument{%
  }

\usepackage{amsmath}
\usepackage{xcolor}
\usepackage{longtable}
\usepackage{array}
\usepackage{afterpage}
\usepackage{booktabs}
\usepackage{tcolorbox}
\usepackage{multirow}
\usepackage{enumitem}

\definecolor{darkgreen}{rgb}{0,0.5,0}

\setcopyright{acmlicensed}
\copyrightyear{2025}
\acmYear{2025}
\acmDOI{XXXXXXX.XXXXXXX}
\acmISBN{978-1-4503-XXXX-X/2025/06}

\begin{document}

\title{Insights on Adversarial Attacks for Tabular Machine Learning via a Systematic Literature Review}

\author{Salijona Dyrmishi}
\email{salijona.dyrmishi@uni.lu}
% \orcid{}
\author{Mohamed Djilani}
\email{mohamed.djilani@uni.lu}
\author{Thibault Simonetto}
\email{thibault.simonetto@uni.lu}
\author{Maxime Cordy }
\email{maxime.cordy@uni.lu}
\affiliation{%
  \institution{University of Luxembourg}
  \country{Luxembourg}
}

\author{Salah Ghamizi}
\affiliation{%
  \institution{Luxembourg Institute of Health}
  % \city{Hekla}
  \country{Luxembourg}}
\email{salah.ghamizi@lih.lu}
\affiliation{%
  \institution{RIKEN AIP}
  % \city{Hekla}
  \country{Japan}}
\renewcommand{\shortauthors}{Dyrmishi et al.}

\begin{abstract}
Adversarial attacks in machine learning have been extensively reviewed in areas like computer vision and NLP, but research on tabular data remains scattered. This paper provides the first systematic literature review focused on adversarial attacks targeting tabular machine learning models. We highlight key trends, categorize attack strategies and analyze how they address practical considerations for real-world applicability. Additionally, we outline current challenges and open research questions. By offering a clear and structured overview, this review aims to guide future efforts in understanding and addressing adversarial vulnerabilities in tabular machine learning.
\end{abstract}





\maketitle

\section{Introduction}
\label{introduction}

Structured tabular data remains widely used in high-stake industries such as healthcare, finance, and cybersecurity, where its standardized format, interpretability, and reliability make it suitable for supporting critical decision-making. Machine learning models trained on tabular data (Tabular ML models) help to diagnose medical conditions, detect financial fraud, and identify network anomalies, enabling more accurate and efficient decision-making. While recent advances in deep learning have mostly focused on unstructured data types such as images and text, there is growing interest in adapting deep learning architectures to tabular domains. This includes the development of models like  TabM \cite{gorishniy2024tabm}, SwitchTab \cite{wu2024switchtab}  or the foundational model TabPFN (\cite{hollmann2025tabpfn}), which aim to combine the strengths of neural networks with the unique characteristics of tabular data. These advancements open new possibilities for leveraging tabular data and achieving performance gains.

Tabular ML models are vulnerable to adversarial attacks, much like their counterparts in other data modalities. These attacks, also known as evasion attacks, involve small, carefully crafted perturbations to input data that manipulate model predictions. They were initially studied in the image domain, where invisible pixel perturbations drastically changed the model's decision. Similar vulnerabilities in tabular settings involve targeted modifications to critical features such as financial transaction amounts, patient lab results, or network traffic metrics, leading to wrong predictions. For instance, a fraud detection model could be deceived into misidentifying fraudulent transactions as legitimate by changing the transaction amount from 100 euros to 101 euros. This vulnerability of Tabular ML models was first brought to attention by \citet{ballet2019imperceptible}, with subsequent research exploring these risks across various tabular applications, including credit scoring \cite{ghamizi2020search}, fraud detection \cite{agarwal2021black}, phishing \cite{gressel2021feature}, and intrusion detection \cite{teuffenbach2020subverting}. 

Unlike adversarial attacks on continuous data like images, those on tabular data present unique challenges due to the inherent heterogeneity and interdependencies in such datasets. Tabular data often includes a mix of numerical and categorical variables, some of which may be immutable (e.g., demographic information like age) or governed by strict logical relationships (e.g., a loan amount should not exceed a certain percentage of income). Crafting adversarial examples that respect these domain-specific constraints while successfully misleading a model is significantly more complex than perturbing pixel values in images. Furthermore, many real-world applications often involve constraints like regulations, business logic, or physical limitations that make many adversarial perturbations infeasible. As a result, directly applying unrestricted attack methods from computer vision often yields unrealistic examples with limited relevance to tabular data models. To meaningfully assess and improve model robustness in practice, it is essential to generate adversarial examples that are not only effective in misleading models but also feasible within the real-world data and domain context. These challenges bring the need for tailored adversarial attack strategies on tabular data that explicitly account for data validity and domain constraints.

Given the important role of tabular ML for many critical applications and its susceptibility to adversarial attacks, it is important to understand, analyze, and mitigate adversarial threats in these models. 
Although the adversarial ML literature for unstructured data such as images \cite{long2022survey}, text \cite{zhang2020adversarial}, and audio \cite{lan2022adversarial}, has matured with several well-established surveys, the corresponding body of work for tabular data remains fragmented. This fragmentation is largely due to the tendency of existing studies to focus on specific application domains rather than addressing tabular adversarial attacks as a unified research area. While a few surveys have been conducted within isolated contexts such as intrusion detection and spam filtering \cite{martins2020adversarial, he2023adversarial, alotaibi2023adversarial, imam2019survey}, there is still no domain-agnostic synthesis that systematically describes attack methodologies and practical considerations for tabular ML. This gap motivates the need for a systematic literature review that consolidates existing research, identifies common trends and challenges, and lays the ground for future work in this field.
 
Therefore, we conduct for the first time in our knowledge a systematic review of the landscape of adversarial attacks on tabular ML models, consolidating insights from a wide range of application domains. Following the PRISMA methodology as detailed in Section \ref{sec:methodology}, we identified and analyzed 53 relevant studies. We begin by examining \textbf{\textit{publication trends}} through metadata such as publication year and venue (Section \ref{sec:RQ1}). Next, we review existing \textbf{\textit{attack methodologies}} tailored to tabular data, categorize them and identify their differences with traditional image-based attacks (Section \ref{sec:RQ2}). We then explore the extent to which these attacks account for eight identified \textbf{\textit{practical considerations}} relevant to real-world applications, including \textit{efficacy, efficiency, transferability, feasibility, semantic preservation, plausibility, defense awareness} and \textit{dataset suitability} (Section \ref{sec:RQ3}). Finally, we discuss current limitations in the literature and suggest promising directions for future research. Through this comprehensive synthesis, we aim to deepen the understanding of adversarial machine learning in the context of tabular data, highlight prevalent attack strategies, and identify critical challenges and research gaps that merit further exploration.

\section{Background}
\label{background}

\subsection{Adversarial Attacks}
\label{subsec:attacks}
Adversarial attacks refer to the deliberate perturbation of input data to induce erroneous predictions from machine learning models (\cite{biggio2013evasion, szegedy2013intriguing}). The most common definition of adversarial examples that are produced by adversarial attacks comes from the image data. Formally, given a classifier \( f: \mathbb{R}^d \to \mathcal{Y} \), an input sample \( x \in \mathbb{R}^d \) with true label \( y \in \mathcal{Y} \), an adversarial example \( x^{\text{adv}} \) satisfies $\|x^{\text{adv}} - x\|_p \leq \epsilon \quad \text{and} \quad f(x^{\text{adv}}) \ne y$
where \( \epsilon > 0 \) is a small constant that controls the magnitude of the perturbation under the \( \ell_p \)-norm. Adversarial attacks are typically categorized based on the adversary's goal, capabilities, and knowledge:

\begin{itemize}
    \item \textbf{Goal}: The primary objective of an adversarial attack is to mislead the model into making an incorrect prediction. Two key types of attacks can be distinguished:
    \begin{itemize}
        \item \textit{Untargeted attack:} The adversary aims for any misclassification, causing the model to predict any incorrect label without specifying a particular target.
        \item \textit{Targeted attack:} The adversary seeks to cause the model to predict a specific incorrect label, which is predefined.
    \end{itemize}
    
    \item \textbf{Knowledge}: The knowledge refers to the information the adversary has about the target model. This knowledge can vary as follows:
    \begin{itemize}
        \item \textit{Complete knowledge (White-box attacks)}: The adversary has full access to the model, including its architecture, parameters, and training data. These attacks represent the worst-case scenario.
        \item \textit{Limited knowledge (Grey-box attacks)}: The adversary knows some aspects of the model but does not have complete access. For instance, they may know the model's architecture but not its parameters, or they may have access to the training dataset.
        \item \textit{No knowledge (Black-box attacks)}: The adversary has no access to the model’s internals and can only observe input-output pairs. Black-box attacks are more common in real-world scenarios. They can be further categorized into:
        \begin{itemize}
            \item \textit{Query-based attacks:} The adversary interacts with the target model by submitting input queries and observing the corresponding model outputs. Based on the feedback received, the adversary iteratively crafts adversarial examples. 
            
            \item \textit{Transfer-based attacks:} These attacks exploit the transferability of adversarial examples. The adversary generates adversarial examples using a surrogate model (which may be a different model than the target model or a model with similar architecture or data) and applies these examples to the target model. The assumption is that adversarial examples crafted for one model can also mislead other models with similar architectures or decision boundaries \cite{papernot2016transferability}.
            \item \textit{Hybrid attacks:} Some attacks combine query-based and transfer-based strategies, where the adversary uses query-based feedback to fine-tune adversarial examples generated using a surrogate model. 
        \end{itemize}
    \end{itemize}
    
    \item \textbf{Capabilities}: The adversary's capabilities refer to the methods available for generating adversarial examples, which depend on the information they have access to. These methods include, but are not limited to, the following strategies:
    \begin{itemize}
        \item \textit{Gradient-based methods:} These attacks rely on the gradients of the loss function with respect to the input. They use the model's gradients to adjust the input data in the direction that maximizes the loss w.r.t to the correct label, thereby inducing wrong predictions. These methods are effective in white-box settings, where the adversary has full access to the model’s parameters, as well as in transfer-based attacks, where adversarial examples generated from a surrogate white-box model are used to target a black-box model.

        \item \textit {Gradient-free methods:}
        These methods rely solely on fitness functions derived from the model's output and do not need internal access to the model to calculate gradients. Depending of the target scenario, such attacks are classified either as score-based attacks, where the attacks leverage the fine-grained output scores (like probabilities) of the model, or decision-based attacks which rely solely on class labels. Gradient-free methods can be split further down, but not limited to, in two categories: 
        \begin{itemize}
            \item \textit{Rule-based methods:} These methods rely on predefined rules and heuristics to generate adversarial perturbations. Rule-based methods are generally simpler and faster to implement, 
            but may be less effective in more complex attack scenarios compared to gradient-based or learning-based methods.

            \item \textit{Meta-heuristic methods:} These methods utilize nature-inspired optimization algorithms to partially explore the input space, guided by a fitness function. 
            In the context of adversarial attacks, the fitness function typically includes misclassification, possibly along with additional goals such as minimizing perturbation magnitude or satisfying soft constraints. Meta-heuristic methods are typically iterative and evaluation-intensive in  nature.
                        
        \end{itemize}

        \item \textit{Learning based Methods:} These approaches involve training a model  to learn a mapping from clean inputs to adversarial examples, allowing for fast, one-step attack generation once trained. While efficient at inference, they require significant training data to perform effectively.
    \end{itemize}
\end{itemize}

\subsection{Practical Considerations for Adversarial Attacks}
\label{subsec:aspects}

Practical considerations play an important role in the development and evaluation of adversarial attacks on tabular ML models. These considerations serve both as performance criteria and as constraints, ensuring that adversarial examples are not only effective in deceiving models but also plausible and applicable in real-world scenarios. While these considerations overlap with those in other data modalities, their application to tabular data requires new perspectives due to the domain's unique characteristics, particularly its mixed feature types and strict inter-feature dependencies. Below, we summarize key practical considerations discussed in the literature on adversarial attacks.

\textbf{Efficacy.}  
It measures how successfully the crafted inputs cause erroneous predictions. In classification tasks, efficacy is often quantified by the \textit{attack success rate} (ASR), which is the percentage of adversarial examples that result in incorrect model outputs. Mathematically, the attack success rate is defined as $\text{ASR} = \frac{N_{\text{misclassified}}}{N_{\text{total}}}$ where $N_{\text{misclassified}}$ is the number of misclassified adversarial examples and $N_{\text{total}}$ is the total number of generated adversarial examples. 
 Complementary to this is \textit{robust accuracy} (RA), which captures the model’s accuracy when subjected to adversarial perturbations. It is defined as $\text{RA} = \frac{N_{\text{correct}}}{N_{\text{total}}}$ where \( N_{\text{correct}} \) is the number of correctly classified adversarial examples, and \( N_{\text{total}} \) is the total number of generated adversarial examples. A lower robust accuracy indicates a model more vulnerable to adversarial examples. By definition, ASR and RA satisfy the relation $\text{RA} = 1 - \text{ASR}$ if and only if the adversarial examples are generated only from inputs that are originally correctly classified.

\textbf{Efficiency.} It describes how resource-intensive an adversarial attack is in terms of computation time, memory usage, and the number of model queries required to generate effective perturbations. Efficiency plays a critical role in real-world environments, where factors like API rate limits, system latency, and detection mechanisms may limit the adversary's ability to interact with the model.

\textbf{Transferability.}  It refers to the ability of adversarial examples crafted for one model to mislead another model, potentially with a different architecture or training data. This property is especially important in black-box attack scenarios, where the adversary can craft adversarial examples for a surrogate model and tests their performance on the target model (\cite{papernot2016transferability}). The effectiveness of transferability can be quantified by the success rate of adversarial examples across different models: $\text{TR} = \frac{N_{\text{misc\_target}}}{N_{\text{total}}}$ where \( N_{\text{misc\_target}} \) is the number of adversarial examples that successfully mislead the target model, and \( N_{\text{total}} \) is the total number of adversarial examples that were able to cause mis-predictions for the source model and applied to the target model.

\textbf{Feasibility.}  It indicates whether perturbed examples remain valid within the structural and semantic rules of the application domain. Unlike image data, where small perturbations generally preserve visual realism, tabular data is more structured, consisting of various feature types such as categorical, ordinal, discrete, and continuous variables. Each feature in tabular data comes with domain-specific constraints that must be respected. Violating these constraints, such as assigning an invalid category, producing an out-of-range numerical value, or breaking feature dependencies, makes the adversarial example unrealistic and easily detectable by manual or automated data quality tools. There are a few practical constraints related to feasibility in tabular data:

\begin{itemize}
    \item \textit{Mutability constraints.} In tabular data, mutability constraints define which features must remain unchanged during adversarial perturbations due to domain-specific or practical limitations. These typically include attributes, such as blood group or date of birth, which are either immutable by nature or prohibitively costly or infeasible for an adversary to manipulate. Mutability constraints are typically handled by masking immutable features, allowing perturbations to be applied only to the remaining modifiable ones.
    
    \item \textit{Structural Constraints.}  Tabular features are typically bound by predefined domains and data types. For instance, age may be restricted to integer values within [0, 120], and categorical features must belong to a fixed set of classes. Likewise, binary indicators and one-hot encoded fields must remain strictly binary. To preserve data integrity, adversarial perturbations must respect these structural constraints. This is often achieved using techniques such as clipping, rounding, or projection to ensure that modified inputs remain within valid value ranges and formats.
    
    \item \textit{Inter-feature Dependencies.} Tabular datasets often contain logical or statistical dependencies between features. These dependencies may include domain-specific rules (e.g., \texttt{LoanAmount} should not exceed a certain percentage of \texttt{Income}) or engineered relationships (e.g., \texttt{TotalDebt} = \texttt{MortgageDebt} + \texttt{CreditCardDebt}). To preserve these dependencies, adversarial attacks must employ methods such as constraint-aware optimization, rule-based validation, or post-hoc projections that ensure consistency across related features.
\end{itemize}

Adversarial perturbations that fail to respect mutability, structural, and inter-feature dependencies compromise data validity and are often easily detected by automated systems or domain experts. As a result, preserving these requirements contributes to ensuring the realism and stealth of adversarial examples in practice. This requirement is emphasized in the TabularBench benchmark (\cite{simonetto2024tabularbench}), which explicitly filters out adversarial examples that violate such constraints. 

\textbf{Semantic preservation.}  
It refers to whether adversarial perturbations preserve the semantic meaning of the input, ensuring that the changes do not distort the real-world interpretation of the data (e.g., a slightly altered image of a cat remains recognizable as a cat). In image tasks, \( L_p \)-norm constraints are commonly used to limit perturbations and ensure that they remain visually imperceptible. This works well because each pixel is a continuous value within a fixed range (e.g., 0–255), and small changes in pixel intensities typically result in a perceptually similar image, preserving the semantics. However, when applied to tabular data, \( L_p \)-norm constraints are less effective. In this case, \( L_p \)-norms measure only numerical proximity and do not guarantee that the perturbations are semantically valid. For example, changing a feature like \texttt{``employment status''} from \texttt{``employed''} to \texttt{``unemployed''} may result in a minimal \( L_0 = 1 \) change but can completely alter the input’s context and outcome. In a loan approval scenario, such a change would likely lead to a rightful rejection, as the applicant now has a higher risk.

\textbf{Plausibility.} It captures whether an adversarial example appears realistic and coherent within the expected distribution of the data, particularly from the perspective of a human observer or domain expert. In the image domain, plausibility is closely related to semantic preservation and imperceptibility, addressed by \( L_p \) norm perturbations. Slight pixel-level modifications, when constrained by an \( L_p \)-norm, are typically imperceptible to humans and do not distort the perceived image. While minimal \( L_p \)-norm perturbations in images can preserve both semantics and plausibility, the same is not directly applicable to tabular data. Minimal perturbations in tabular data may (or may not) preserve semantic meaning, but when they fail to adhere to the logical or business rules that domain experts expect, they are implausible. For example, in a credit scoring dataset, slightly increasing an applicant’s income might not change their overall risk category. However, if the reported income becomes inconsistent with other features, such as employment status,  or declared expenses, it could raise red flags during manual or automated inspections. 
% For example, in a medical dataset, altering a patient's blood pressure by a minimal amount might preserve the overall health diagnosis category. However, if the change is not consistent with other features like age or medical history, it could raise concerns with a domain expert. 

% For example, in a credit scoring scenario, a small perturbation to an individual's annual income while keeping employment status unchanged may preserve the semantics (e.g., income range remains realistic). However, it could still be implausible to a domain expert if the income is slightly below the average for similar occupations. 

\textbf{Defense awareness.} It considers whether an adversarial attack accounts for the presence of defenses in the target model. In practice, models may employ defenses such as adversarial training \cite{madry2017towards}, feature masking \cite{gao2017deepcloak}, or input sanitization \cite{xie2019feature}. Evaluating attacks only on undefended models overlooks the impact of such defenses on model behavior. To ensure realistic assessments, attacks should also be tested against defended models, requiring the development of adaptive attacks that consider the specific defenses in place. As new defenses emerge, testing on both defended and undefended models is necessary for a comprehensive evaluation of adversarial robustness in tabular ML.

\textbf{Dataset suitability.} It reflects how well the evaluation dataset captures real-world conditions, and whether its selection is justified and appropriate for the attack scenario. Using a limited number, overly simplistic, synthetic, or narrowly scoped datasets can reduce the relevance of the results.Suitable datasets exhibit diversity, are aligned with real application domains, and support well-defined threat models. Justifying dataset selection and acknowledging its limitations are essential for ensuring that attacks are not overly tailored to specific benchmarks and that results translate to practical settings.

\section{Methodology}
\label{sec:methodology}

\begin{figure}[ht!]
    \centering
    \includegraphics[width=\textwidth, trim=0cm 1.8cm 0cm 1.8cm, clip]{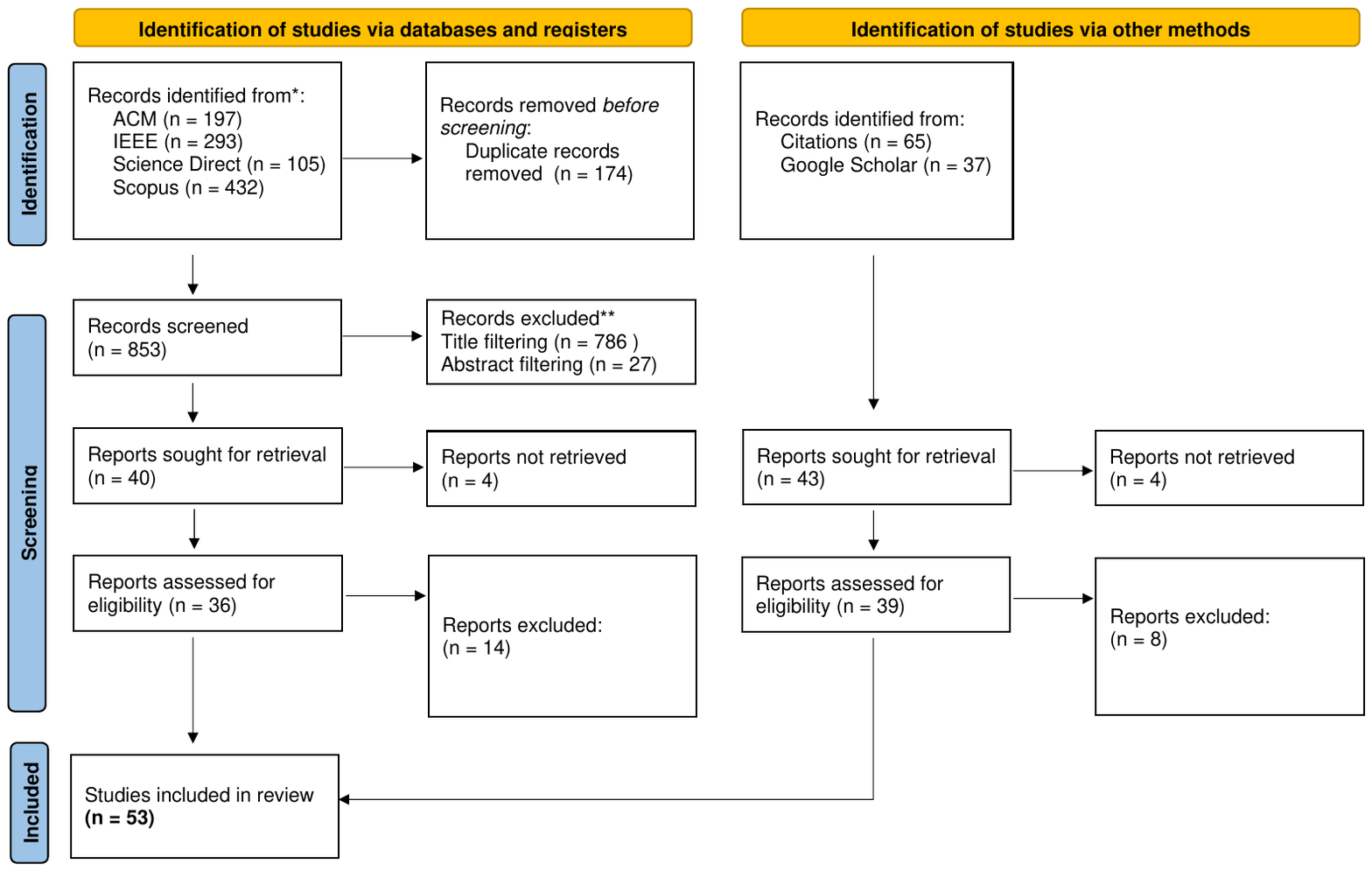}
    \caption{Overview of the research paper screening process.}
    \label{fig:prisma_overview}
\end{figure}

This study follows the Systematic Literature Review (SLR) methodology ~\cite{kitchenham2004procedures}, ensuring a structured and reproducible approach to identifying and analyzing adversarial attacks on tabular data. To enhance transparency, we adhere throughout the selection and evaluation process to the PRISMA 2020 guidelines  \cite{page2021prisma}.

\textbf{Aim and Scope.} This systematic review aims to comprehensively examine the current landscape of adversarial attacks targeting tabular ML models. To our knowledge, this is the first systematic effort to consolidate and analyze findings from a wide range of domains specifically concerning tabular data. We identified and reviewed 53 academic publications written in English that introduce or apply adversarial attack techniques to tabular models. This review excludes works focused primarily on defense strategies, robustness benchmarking using established attacks, or methods based on random perturbations without a principled adversarial formulation. Our scope includes analyzing publication trends in adversarial attack research for tabular data, organizing and synthesizing the methodologies employed across different studies, and evaluating the extent to which these attacks address practical real-world considerations, such as those defined in Section~\ref{subsec:aspects}. Through this structured analysis, we aim to discover methodological patterns, highlight overlooked challenges, and identify possible future research directions. The review is guided by a set of research questions outlined in Section~\ref{rqs}.

\textbf{Data Collection.} The data collection process took place between February 24, 2025, and March 4, 2025, across four major academic databases: IEEE Xplore, Scopus, ACM Digital Library, and Semantic Scholar. The search query used was designed to broadly capture studies that propose novel adversarial attack methodologies for tabular data while filtering out surveys and review articles. The following search string was applied, with adaptations made for each database’s syntax: \texttt{AllField:(("adversarial attack*" OR "adversarial example*") AND "tabular data") AND Title:(NOT (survey OR review))}. After merging search results and removing duplicates, an initial list of 853 papers was identified.

\textbf{Study Selection and Filtering.} A multi-stage filtering process was implemented to refine the selection. Initially, papers were screened based on their titles to eliminate irrelevant studies. This was followed by an abstract-based filtering stage, where papers that did not explicitly propose novel adversarial attacks for tabular data were removed. After this stage, 40 papers remained and were subjected to full-text retrieval; however, 4 of these were inaccessible due to paywall restrictions. The remaining 36 papers underwent a thorough full-text assessment to determine their eligibility according to predefined inclusion and exclusion criteria.  Studies were included if they: \textit{a)} were written in English and \textit{b)} introduced a new adversarial attack technique specifically targeting tabular ML. Excluded studies were those that: \textit{a)} primarily focused on defense mechanisms, \textit{b)} evaluated robustness using previously established adversarial attacks, \textit{c)} employed random perturbation-based strategies rather than systematic adversarial methodologies, or \textit{ d)} operated before feature extraction into a tabular form.
In the end, 22 studies met these criteria and were included in the review.

\textbf{Additional Sources.} 
To expand our set of relevant studies, we used a snowballing approach by identifying articles that cited or were cited by the 22 studies identified in the initial screening. This yielded 65 additional articles. To further complement the database results, we also conducted a manual search on Google Scholar using the same search terms as in the primary search. Due to the limitations of automated extraction, we limited this search to the top 100 results ranked by Google Scholar’s relevance algorithm, which added 37 more articles. These 102 papers were screened using the same process described in the \textit{“Study Selection and Filtering”} section above. After title and abstract screening, 43 papers were selected for full-text retrieval. Of these, 39 were successfully retrieved and assessed, of which 31 met the inclusion criteria.

Combining the selected papers from the initial set and the additional resources, we retained in total \textbf{53 papers} for detailed analysis. An overview of the selection process is presented in  Figure~\ref{fig:prisma_overview} using the PRISMA 2020 flowchart, which shows the number of records identified, screened, excluded, and included in the final dataset. 

\textbf{Data Extraction.} For each selected article, we extracted metadata and relevant research information. The metadata included author names, paper title, year of publication, publication venue, citation count, and the main research focus.
For the research-relevant content, we analyzed the adversarial threat model, including the adversarial goal, knowledge level, and capabilities. We also collected information on the paper’s awareness of practical considerations outlined in Section~\ref{subsec:aspects} necessary for real-world applicability. The extracted data were synthesized to provide an overview of the current landscape of adversarial attacks for tabular ML, highlighting emerging trends, common challenges, and potential directions for future work.
\section{Research Questions}
\label{rqs}

We structure our study around the following research questions to systematically analyze the state of research on adversarial attacks for tabular ML models.  

\textbf{RQ1: \textit{What are the key research trends and evolution patterns in adversarial attacks on tabular ML models?}} 

Understanding the historical progression and current trends helps contextualize the state of the field. This question investigates the evolution of research in this domain by analyzing publication trends over time, including the number of studies published annually and the venues (conferences and journals) where they appear. We also identify dominant themes in the literature and assess citation metrics to highlight influential studies and methodologies that have shaped the field.

\textbf{RQ2: \textit{Which methods are currently used to generate adversarial examples for tabular ML models?}} 

This question examines the range of techniques developed to craft adversarial examples for tabular data. We collect and report details of the corresponding threat models, including adversarial goals, knowledge, and capabilities. 
We analyze whether these methods adapt attacks originally developed for image-based models or are specifically designed for tabular data. In addition, we investigate how each method addresses the inherent characteristics of tabular features.

\textbf{RQ3: \textit{How do the reviewed 
 works address the practical considerations for adversarial attacks on tabular ML models?}}
 
 This question examines how the attack methods reviewed in RQ2 incorporate the eight key aspects of practical adversarial attack design and evaluation  (efficacy, efficiency, transferability, feasibility, semantic preservation, plausibility, defense awareness, dataset suitability). We assess which of these factors are explicitly addressed, the degree of emphasis placed on each, and how these choices reflect the readiness of the methods for deployment in real-world applications. This helps reveal whether existing research aligns with practical concerns or remains largely theoretical.

By addressing these research questions, this study aims to provide a comprehensive overview of adversarial attacks on tabular ML, highlight gaps in the current research landscape, and identify promising directions for future work.

\section{Research Findings}
\label{results}
This section presents the key findings derived from the analysis of meta-data and research content extracted from the 53 selected studies. The analysis addresses the three research questions presented earlier, providing insight into research trends, adversarial methods, and practical considerations.

\subsection{ RQ1: Research Trends and Advances}
\label{sec:RQ1}

We begin by analyzing research trends through publication timelines, venue distribution, research focus, and citation patterns. The complete extracted metadata used for this analysis is provided in Appendix \ref{app:rq1}.

\begin{figure}[ht!]
    \centering
    \begin{minipage}{0.49\textwidth}
        \centering
        \includegraphics[width=\linewidth]{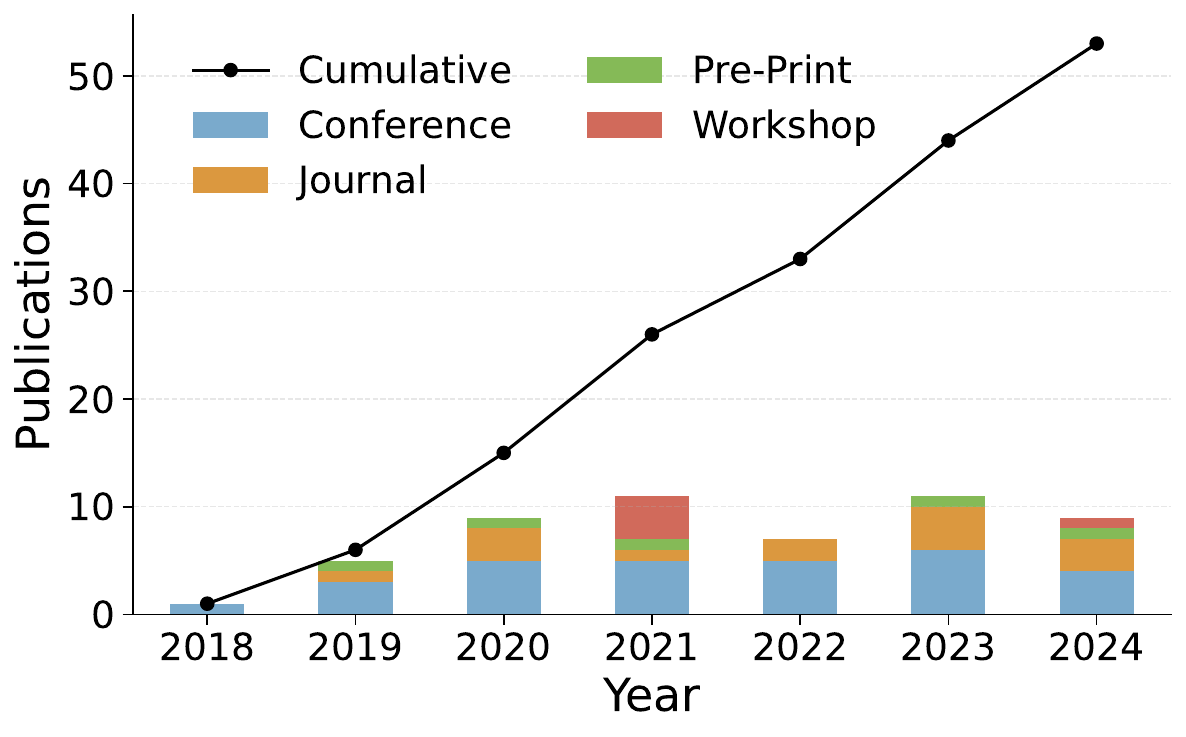}
        \caption{Publications over time by venue type.}
        \label{fig:pub_years}
    \end{minipage}
    \hfill
    \begin{minipage}{0.49\textwidth}
        \centering
        \includegraphics[width=0.89\linewidth]{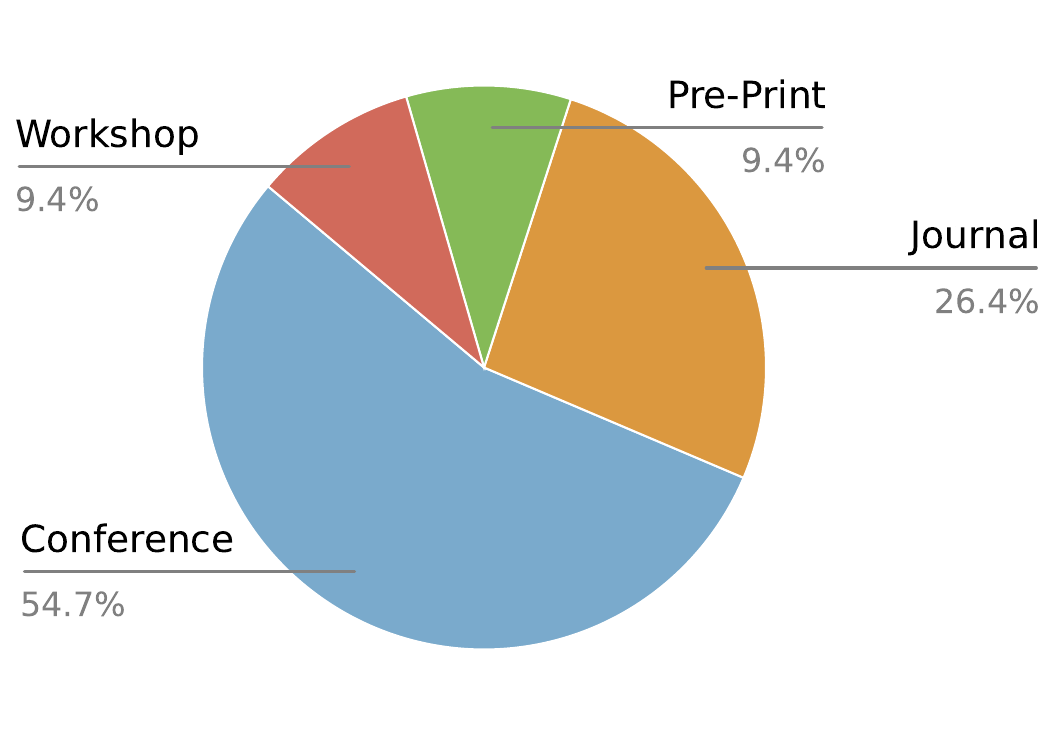}
        \caption{Publications by venue type.}
        \label{fig:pub_venue}
    \end{minipage}
\end{figure}

\textbf{Publication Timelines.} Figure~\ref{fig:pub_years} presents the annual publication counts from 2018 to 2024\footnote{The data reported here is based on the year of first preprint appearance, when applicable, as it better reflects when the work became publicly available. Exact publication years in peer-reviewed venues are listed in Appendix \ref{app:rq1}.}, split by type of venue. The earliest identified use of adversarial attacks on tabular data appears in \cite{lin2022idsgan}, which targeted network intrusion detection systems (NIDS). Although it employed tabular inputs, the study did not frame the task as a general adversarial attack on tabular data. In contrast, \cite{ballet2019imperceptible} is widely recognized as the first domain-agnostic work, establishing general principles for adversarial attacks in tabular settings. From 2018 to 2021, the number of publications increased steadily, reflecting growing research interest. This trend culminated in a publication peak in 2021, marked by several key contributions \cite{gressel2021feature, erdemir2021adversarial, sheatsley2021robustness,simonetto2021unified} that introduced novel attack strategies, addressed domain-specific constraints, and deepened theoretical insights into robustness for tabular ML. After 2021, the number of publications began to stabilize. This transition aligns with broader trends in machine learning security, including rising interest in foundation models and generative AI threats. These developments may have contributed to a relative decline in research in later years, specifically addressing adversarial robustness in tabular ML.

\textbf{Publications by Venue.} We examine how publications are distributed by venue type, as seen historically in Figure \ref{fig:pub_years} and overall in Figure \ref{fig:pub_venue}. We further detail their distribution according to venue specialties in Table \ref{tab:pub_venues_by_type}. Conferences are the most frequent venues with 54.7\% of publications. This reflects the fast-moving pace of the field, where new methods benefit from quick peer feedback. Well-known conferences include IEEE S\&P, NDSS, AAAI, NeurIPS, and ICML. Journals make up 26.4\%, primarily from the cybersecurity space such as JCS, TOPS, and TDSC. Notably, none of the reviewed papers were published in AI or ML journals. Workshops and pre-prints each account for 9.4\%. While often used for early-stage work, some of the most cited and impactful contributions in this area \cite{ballet2019imperceptible, sheatsley2021robustness, gressel2021feature} are still only available as preprints. Regarding the distribution over time, Figure \ref{fig:pub_years} shows an increase in both conference and journal publications. Notably, the ratio between journals and conferences between 2022 and 2024 remains stable, while a decrease of conferences compared to journals would have indicated a trend towards slower and more mature work. 

The majority of publication venues fall within the Cybersecurity and AI \& ML domains, with 14 papers each (ref. Table \ref{tab:pub_venues_by_type}). Other areas include Big Data, Data Mining, and Knowledge Discovery (8 papers), System Design \& Engineering (5 papers), and Computer Networks \& Communication (3 papers). An additional 9 publications are spread across various other venues. An interesting observation is that research on adversarial attacks for tabular data is highly fragmented, with nearly as many venues as publications (46 vs 53). This reflects its interdisciplinary nature, which spans cybersecurity, artificial intelligence, and domain-specific applications. It also highlights the need for more unified benchmarks and stronger interdisciplinary collaboration.

\begin{table}[ht!]
\footnotesize
\centering
\caption{Publication Venues by Domain and Venue Type.}
\label{tab:pub_venues_by_type}
\renewcommand{\arraystretch}{1.2}
\begin{tabular}{@{}p{4cm}p{1.5cm}p{5cm}p{0.8cm}p{0.8cm}@{}}
\toprule
\textbf{Domain} & \textbf{Venue Type} & \textbf{Venues} & \textbf{\#Ven} & \textbf{\#Pub} \\
\midrule

\multirow{2}{=}{Cybersecurity} 
& Journal & JCS, TOPS, JISA, DTRAP, C\&S & 5 & 7 \\
& Conference & DBSec, ARES, SP,  AsiaCCS, CCS, NDSS, DSC & 7 & 7 \\

\cmidrule(r){1-1}

\multirow{3}{=}{AI \& ML}

& Conference & MAKE, NeurIPS, ICMLA, IJCAI, AAAI, IJCNN, ICML, ICMLC, CVIDL & 9 & 10 \\
& Workshop & AI Safety, SafeAI, NextGenAISafety,  ML4Cyber & 4 & 4 \\
\cmidrule(r){1-1}
\multirow{3}{=}{Big Data, Data Mining, \& Knowledge Discovery}
& Journal & KBS, TKDD,  & 2 & 2 \\
& Conference & PAKDD, KDD, CIKM, ITADATA, BigDIA & 5 & 5 \\
& Workshop & MUFin21 & 1 & 1 \\
\cmidrule(r){1-1}
\multirow{2}{=}{System Design \& Engineering}
& Journal & ESWA, TDSC, SMC & 3 & 3 \\
& Conference & USBEREIT, ICPADS & 2 & 2 \\
\cmidrule(r){1-1}
\multirow{2}{=}{Computer Networks \& Communication}
& Journal & Future Internet, TNSM & 2 & 2 \\
& Conference & IWCMC & 1 & 1 \\
\cmidrule(r){1-1}

\multirow{2}{=}{Others}
& Conference &  ICAIF, ESEC/FSE, IIKI, IPSA & 4 & 4 \\
& Pre-print &  arXiV & 1 & 5 \\
\bottomrule
\end{tabular}
\end{table}

\begin{figure}[t!]
    \centering
    \begin{minipage}{0.49\textwidth}
        \centering
        \includegraphics[width=\linewidth]{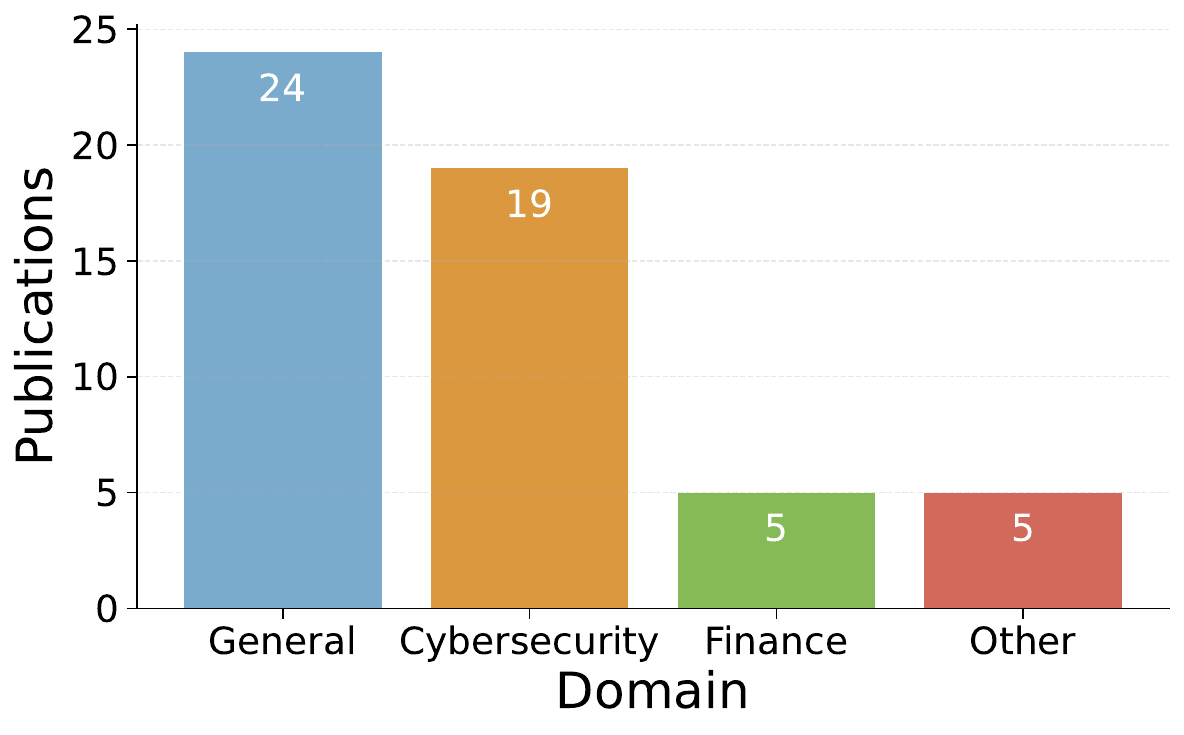}
        \caption{Distribution of research focus by application domains.}
        \label{fig:domain_focus}
    \end{minipage}
    \hfill
    \begin{minipage}{0.49\textwidth}
        \centering
        \includegraphics[width=0.998\linewidth]{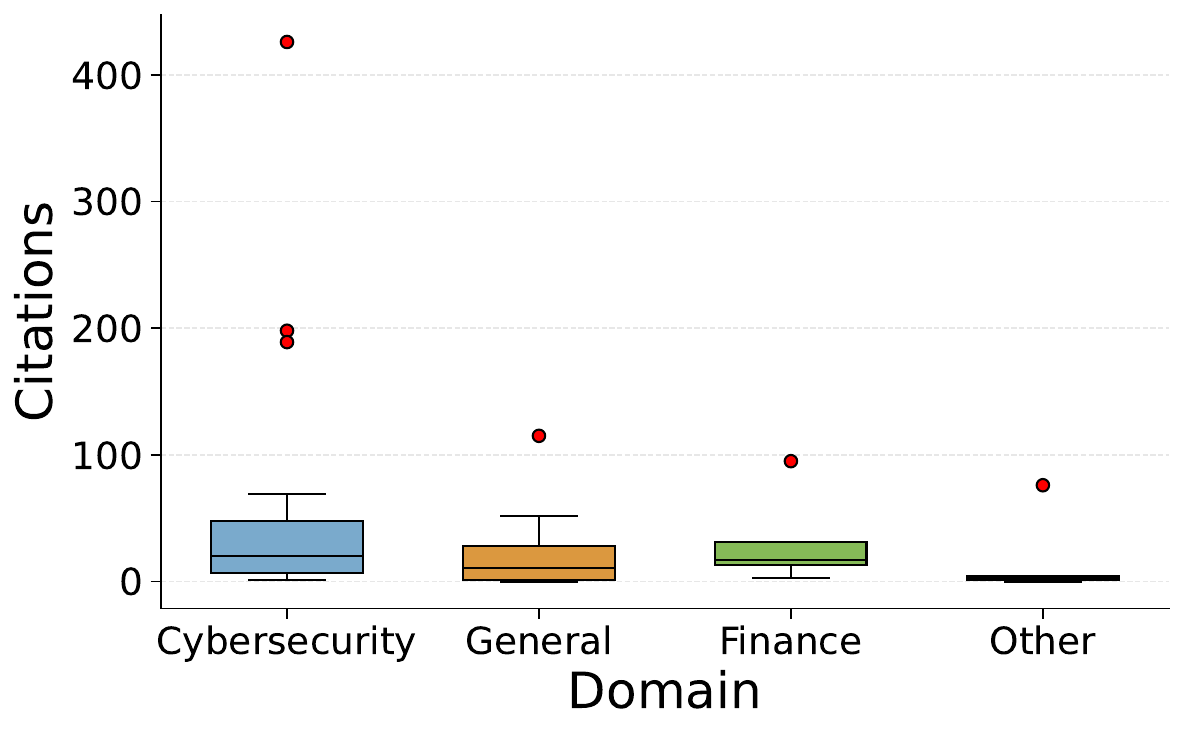}
        \caption{Citations distribution per application domain.}
        \label{fig:citations}
    \end{minipage}
\end{figure}

\textbf{Research Focus.} We analyzed the titles and abstracts of the selected articles to determine the primary research focus as claimed by the authors. The analysis revealed two main directions: general-purpose approaches and domain-specific studies.  Some papers proposed methods claiming to be applicable across a wide range of tabular datasets, while others targeted specific applications in specific domains. \footnote{This distinction does not imply that domain-specific methods cannot be generalized; rather, we categorize the studies based on how the authors themselves position their work.}   Figure \ref{fig:domain_focus} shows that 24 publications focus on general tabular data, 19 on cybersecurity applications (NIDS, phishing, spam), 5 on financial applications (fraud, credit scoring), and 5 on other application domains (CPS, IoT, click-through rate, access control). This breakdown highlights the interest in domain-specific adversarial attack strategies that account for particular domain constraints.  Given the inherently adversarial nature of cybersecurity, where robustness and attack resilience are critical, it is unsurprising that it accounts for 36\% of the studies.

In addition to domain-specific goals, many studies focus on methodological aspects of adversarial attacks on tabular data and practical concerns for real-world implementation. Several articles address components of the threat model, with 6 studies exploring black-box scenarios with limited knowledge (e.g., \cite{agarwal2021black}) and 12 examining adversarial capabilities, including the design of novel attack strategies (e.g., \cite{gressel2021feature}), targeted attacks (e.g., \cite{parfenov_investigation_2023}) and robustness considerations for specific data types, such as regression tasks and categorical features (e.g., \cite{gupta_adversarial_2021}). Other papers address practical considerations for real-world applications, with feasibility being the most frequently discussed theme, with 23 papers examining whether adversarial examples can be realistically generated in real-world settings (e.g., \cite{tian2020exploring}). Efficiency appears in four studies, focusing on the computational cost and optimization trade-offs of attack and defense mechanisms e.g (\cite{malik_tetraa_2023}) and only one paper explicitly examines plausibility\cite{ballet2019imperceptible}. These focus areas often overlap, and one paper can address multiple aspects simultaneously, highlighting the multidimensional challenges in developing robust adversarial techniques for tabular data.

\textbf{Citation Trends.} The analysis of citation distributions across application domains (Figure \ref{fig:citations} and Table \ref{tab:citations}) provides insight into the visibility and impact of adversarial attack research on tabular data. The cybersecurity domain tends to receive the highest number of citations. The most cited paper \cite{lin2022idsgan} has accumulated 426 citations since its initial release, while other highly cited works in the top 25\% citation bracket range between 50 and 198 citations. In contrast, the general tabular data category shows a more even citation spread. The most cited paper - the first recognized adversarial attack\cite{ballet2019imperceptible} tailored to tabular ML - has 115 citations, with others ranging between 30 and 52. This suggests a consistent but lower overall impact.  Adversarial attack research in finance also demonstrates notable citation counts, with \cite{cartella_adversarial_2021} reaching 95 citations, indicating interest in adversarial robustness within financial applications. In the "Other" category, only one paper exceeds 70 citations, with a sharp drop-off to just 4 for the next most cited. 

\begin{table}[ht!]
    \centering
    \caption{Top 25\% most cited papers per application domain.}
    \label{tab:citations}
    \begin{tabular}{ll ll ll ll}
        \toprule
        \multicolumn{2}{c}{\textbf{General}} & 
        \multicolumn{2}{c}{\textbf{Cybersecurity}} & 
        \multicolumn{2}{c}{\textbf{Finance}} & 
        \multicolumn{2}{c}{\textbf{Other}} \\
        \cmidrule(lr){1-2} \cmidrule(lr){3-4} \cmidrule(lr){5-6} \cmidrule(lr){7-8}
        Ref & Citations & Ref & Citations & Ref & Citations & Ref & Citations \\
        \midrule
        \cite{ballet2019imperceptible} & 115 & \cite{lin2022idsgan} & 426 & \cite{cartella_adversarial_2021} & 95 & \cite{li2021conaml} & 76 \\
        \cite{chernikova_fence_2022} & 52 & \cite{usama2019generative} & 198 & \cite{ghamizi2020search} & 31 & \cite{duan_attacking_2024} & 4 \\
        \cite{sheatsley2021robustness} & 35 & \cite{alhajjar2021adversarial} & 189 & & & & \\
        \cite{erdemir2021adversarial} & 34 & \cite{chen2020generating} & 69 & & & & \\
        \cite{mathov_not_2022} & 32 & \cite{zhao2021attackgan} & 50 & & & & \\
        \cite{gupta_adversarial_2021} & 30 & & & & & & \\
        \bottomrule
    \end{tabular}
\end{table}

The larger volume and higher citation spread for cybersecurity studies, compared to  studies promoted as general, may be explained by the real-world threats that adversarial attacks pose to high-stakes applications in cybersecurity. Alternatively, it may indicate that general methods are not sufficiently tailored to capture the specific characteristics of realizable attacks in cybersecurity. A final interesting observation is that the most cited works were all first made available between 2018 and 2021. This timing coincides with the publication peak shown in Figure \ref{fig:pub_years}. The high citation counts may partially reflect the advantage of time, but they also suggest that many foundational contributions to this field originated during that window.

\begin{tcolorbox}[
colframe=black,          % Dark border for contrast
colback=gray!5,         % Light gray background (soft & elegant)
coltitle=white,         % White title text
colbacktitle=purple!50!black, % Deep purple title background
title=\bfseries Research Trends and Advances, center title, % Bold title text
sharp corners=south,    % Sharp bottom corners
fonttitle=\bfseries \Large    % Bold title
]
From 2019 to 2021, research on adversarial attacks for tabular data increased steadily before stabilizing in recent years. No common publication venue has emerged during these years, with almost as many venues as papers published (46 vs. 53). Researchers typically frame their work either as domain-specific (mainly in cybersecurity and finance) or as general-purpose attacks for tabular data. From these, cybersecurity-focused papers have  larger citation counts.
\end{tcolorbox}
\subsection{RQ2: Adversarial attack methods for tabular data}
\label{sec:RQ2}

The studies included in this review revealed 61 unique attacks. They vary in terms of their goals (targeted vs. untargeted), attacker knowledge (white-box, gray-box, black-box), and optimization strategies (gradient-based, gradient-free, learning-based, or hybrid approaches). While the goals and threat models largely align with those observed in attacks on other data modalities (e.g., images or text), the optimization strategies in the tabular domain require special consideration due to the structured nature of the data and inter-feature dependencies. Given these domain-specific constraints, we 
 analyze these attacks primarily based on their underlying optimization techniques. Each category is introduced in detail, with corresponding summary tables reporting the associated attack goals and knowledge assumptions. Some papers propose multiple attacks; in such cases, we report the results for each attack separately.

We summarize our findings in Figure \ref{fig:rq2_summary}.  The results indicate that  classification is a heavily studied \textit{adversarial task} (58 out of 61 attacks). 
While classification attacks can theoretically be adapted to regression problems by modifying the objective function, the current research landscape lacks extensions of theoretical frameworks and evaluation strategies for regression settings. This gap may be attributed to several factors: the limited availability of regression datasets, the relatively lower occurrence of adversarial threats in regression scenarios, and the overall simplicity of classification tasks compared to regression. 
A similar imbalance can be seen within classification tasks, where 40 out of 53 focus solely on binary classification, and assume a straightforward generalization to multi-class.

Regarding the \textit{target strategies}, a large portion of the attacks (38) are proposed only in the non-targeted setting. We hypothesize that the non-targeted scenario (1) is easier to implement and study, (2) is equivalent to the targeted setting in binary decision scenarios, and (3) aligns with the minimal goal of attackers e.g. causing any failure of the system. Despite this, there are 14 attacks proposed specifically for the targeted setting, and an additional 9 that support both targeted and non-targeted modes. Meanwhile, the distribution of attacks by \textit{knowledge levels} highlights a preference for black-box compared to white-box attacks {(32 vs 23), and 6 of them being gray-box. This likely reflects practical considerations, where researchers assume limited access to the internal parameters of a model. Finally, results show a relatively balanced use of attack optimization strategies, with gradient-free methods (23) slightly more common than gradient-based ones (19), possibly reflecting a modest preference for approaches that don’t rely on model internals. Learning-based attacks, though fewer (15), reflect a need for data driven efficient approaches, while hybrid methods (6) remain less explored, suggesting potential for future research that combines strengths across techniques.

\begin{figure}[t!]
    \centering        \includegraphics[width=0.7\linewidth]{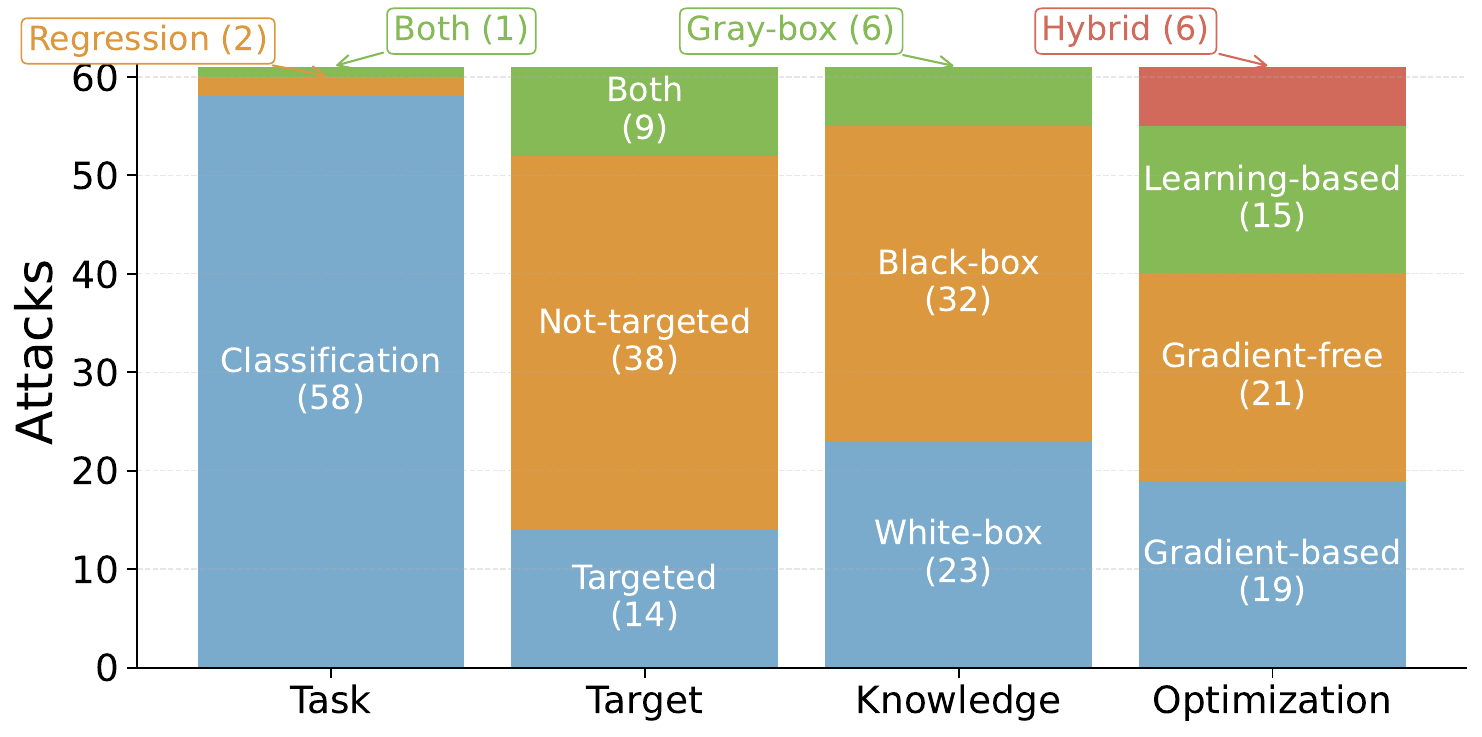}
        \caption{Distribution of attacks across tasks, target, knowledge, and optimization.}
        \label{fig:rq2_summary}
\end{figure}

With the general categorization and summary of attacks established, we now proceed to a detailed examination of each attack category, organized by their underlying optimization strategies.

\subsubsection{Gradient-based approaches}

Table~\ref{tab:gradient_attacks} summarizes the key properties of  gradient attacks for tabular ML.  Most of these attacks are derived from or inspired by classical image attacks, extended to support the characteristics of tabular data. Some others are novel approaches that, while inspired by gradient-based attacks on images, are not direct adaptations of these methods.

\textbf{Adaptations of traditional image attacks}  can be organized into four thematic families based on the original attack  they derive from: Carlini \& Wagner (C\&W) \cite{carlini2017towards}, Iterative Fast Gradient Method (I‐FGSM/BIM) \cite{kurakin2018adversarial} and its improved version Projected Gradient Descent (PGD) \cite{madry2017towards} and Jacobian Saliency Map Attack (JSMA) 
\cite{papernot2016limitations}.

 \begin{itemize}
 \item \textit{C\&W adaptations:}  LowProofFool \cite{ballet2019imperceptible} attack modifies the C\&W loss by adding a feature‐importance regularizer that penalizes changes to the most correlated features with the target, while \citet{nobi_adversarial_2022} extends it to enforce both mutability ranges and one‐hot consistency for mixed categorical and continuous features.  \citet{teuffenbach2020subverting} further guides the C\&W optimizer toward easily altered features by weighting each variable according to its difficulty of modification. Meanwhile, \citet{alhussien2024constraining}  takes a different approach by introducing an explicit projection step after C\&W (and DeepFool \cite{moosavi2016deepfool}) to revert adversarial examples into the valid subspace defined by type, range, and relational constraints.

\item \textit{I-FGSM adaptations:} Constraint‐based I‐FGSM \cite{tian2020exploring} multiplies the input gradient by a binary constraint matrix at each iteration to preserve feature correlations, while \citet{kong2023adversarial} adapts I‐FGSM for regression tasks by maximizing output change within a perturbation budget and scaling each gradient step accordingly. 

\item \textit{PGD adaptations:}  \citet{erdemir2021adversarial} proposes to adjust gradients based on data correlations with the target variable, and \citet{sheatsley2021robustness} combines PGD with a DPLL‐based SAT solver that projects perturbed examples into regions defined by learned CNF constraints, while minimizing the number of perturbed features. Similarly, CaFa \cite{ben-tov_cafa_2024} mines denial constraints via FastADC and enforces them via SAT projection, also bounding the number of modified features. CAPGD \cite{simonetto2021unified,simonetto2024towards} embeds differentiable constraints within the loss and uses constraints repair for equality constraints, adding adaptive step‐size, momentum, and random restarts to improve convergence under these complex restrictions.

\item \textit{JSMA adaptations:}  AJSMA \cite{sheatsley2020adversarial} classifies features as primary or secondary, dynamically filtering out perturbations that violate protocol or relational constraints, whereas \citet{mathov_not_2022}’s latent JSMA first embeds heterogeneous inputs into a continuous space, applies JSMA with Adam‐based updates, and then projects the results back into the valid feature domains.
\end{itemize}

\textbf{Other attacks }
depart from direct image analogs to address domain‐specific dependencies. ConAML~\cite{li2021conaml} constructs a dependency matrix for Cyber‐Physical System variables, backpropagating through independent features and linearly adjusting dependents.\citet{ju_robust_2022} aims to preserve the data distribution by enforcing a KL divergence constraint while also controlling skewness and kurtosis coefficients. FENCE \cite{chernikova_fence_2022} groups features into correlated families, updating both the maximal‐gradient feature and its dependents to maintain semantic consistency. Gupta et al.~\cite{gupta_adversarial_2021} introduce a two‐step process where the Jacobian estimates feature influence before a projected gradient refinement, PCAA \cite{xu2023probabilistic}  models categorical perturbations as distributions, estimating expected gradients and thresholding changes in cross‐entropy loss. Finally, Nandy et al.~\cite{nandy_non-uniform_2023} tackle one‐hot encoded spaces by relaxing them into a continuous domain via reparameterization, then minimizing a composite loss that balances classification error, boundary proximity, entropy regularization, feature correlation, and edit counts.

\begin{table}[t!]
\centering
\caption{Overview of Gradient-based Adversarial Attacks on Tabular Data.}
\label{tab:gradient_attacks}
\footnotesize
\begin{tabular}{@{}p{0.5cm}p{0.5cm}p{1.2cm}p{1.5cm}p{1cm}p{1.5 cm}p{2.8cm}p{1.5cm}p{0.8cm}@{}}
\toprule
\textbf{Ref.} & \textbf{Year} &\textbf{Attack} & \textbf{Task} & \textbf{Targeted} & \textbf{Knowledge} &  \textbf{Description} & \textbf{Domain}&\textbf{Code} \\ \midrule
\cite{ballet2019imperceptible} & 2019 & LowProFool &  Classification & Yes & White-box &  Feat.-weighted C\&W & General  &Yes \\
\cite{chernikova_fence_2022} & 2019 & FENCE & Classification& No & White-box &  Constrained  optimization & General &Yes \\
\cite{tian2020exploring} & 2020 & C-IFGSM & Classification& No & White-box & Constrained I-FGSM & General &No \\
\cite{sheatsley2020adversarial} & 2020 & AJSMA & Classification& Yes & White-box & Constrained JSMA variant  & General & No \\
\cite{teuffenbach2020subverting} & 2020 &-- & Classification& Yes & White-box  & Feat.-weighted C\&W & Cybersecurity& No \\
\cite{mathov_not_2022} & 2020 &-- & Class. \& Reg. & No & White-box & Constrained  JSMA & General & Yes \\
\cite{li2021conaml} & 2020 & -- & Classification& No & White-box  & Best effort search + gr. & Other & No \\
\cite{sheatsley2021robustness} & 2021 & CSP & Classification & No & White-box &  PGD \& JSMA comb. & General & No \\
\cite{erdemir2021adversarial} & 2021 & -- & Classification& No & White-box & Feat.-weighted PGD & General &Yes \\
\cite{gupta_adversarial_2021}& 2021 &  -- & Regression & No & White-box  & Jacobian guided & General & No \\
\cite{ju_robust_2022} & 2022 & -- & Classification& No & White-box  & Gradient optimization & General & No \\
\cite{nobi_adversarial_2022} & 2022& -- & Classification& Yes \& No  & White-box & LowProFool variant & Other & Yes \\
\cite{xu2023probabilistic}& 2022 & PCAA & Classification& No & White-box &  Probabilistic cat. gradient & General & No \\
\cite{kong2023adversarial}&2023 & -- & Regression & No & White-box &  I-FGSM variant & General & No \\
\cite{nandy_non-uniform_2023} & 2023 & DNA & Classification& No & White-box & C\&W version for cat. data & General & No \\
\cite{ben-tov_cafa_2024} & 2024 & CaFA & Classification& No & White-box &  Constrained PGD variant  & General & Yes \\
\cite{simonetto2024towards}&2024 & CAPGD & Classification& No & White-box & Constrained PGD variant & General &Yes\\
\cite{alhussien2024constraining}& 2024 & -- & Classification& Yes \& No  & White-box & Constrained C\&W & Cybersecurity &Yes
\\
\cite{alhussien2024constraining}&2024 & -- & Classification& Yes \& No  & White-box &  Constrained DeepFool & Cybersecurity &Yes
\\
\bottomrule
\end{tabular}
\label{tab:adv_attacks_tabular}
\end{table}

%%%%%%%%%%%%%%%%%%%%%%%%%%%%%%%%%%%%%%%%%%%%%%%%%%%
%%%%%%%%%%%%%%%%%%%%%%%%%%%%%%%%%%%%%%%%%%%%%%%%%%%%%%%%%%%%%%%%%%%%%%%%%%%%%%%%%%%%%%%%%%%%%%%%%%%%%%

Our analysis shows that most gradient-based attacks on tabular data (13 out of 19) are adaptations of classical image-based methods, adjusted to handle tabular-specific constraints like feature mutability, structural consistency, and relationships between features. These adjustments usually involve adding projection steps or embedding constraint penalties directly into the loss function. This means that even though tabular data comes with unique challenges, small changes to existing attacks might be enough in many situations. At the same time, these adaptations can add complexity and require more computation, reflecting a balance between making attacks realistic and keeping them efficient. Because tabular data comes from many different areas (e.g cyber-physical system) some domain-specific constraints and data types may still need custom attack strategies. In particular, handling categorical features and complex relational dependencies often calls for specialized approaches. We also see promising work combining gradient-based methods with symbolic reasoning, projection or constraint solvers, which helps manage these constraints more effectively.

\subsubsection{Gradient-free approaches}

Gradient-free attacks  observed in the reviewed studies (ref. Table \ref{tab:gradient_free_attacks}) fall into three broad families: rule-based methods that rely on predefined heuristics or domain insights, meta-heuristic search strategies that treat adversarial generation as a global black-box optimization problem, and other approaches. 

\textbf{Rule-based attacks} can themselves be subdivided into three core approaches.

\begin{itemize}
 \item \textit{Combinatorial feature exploration} techniques that exhaustively or selectively substitute discrete feature values and enumerate possible adversarial examples. Early work by \citet{shirazi2019adversarial} picks subsets of phishing‐related features and replaces them with values sampled from other phishing examples to find variants that bypass detectors. A later extension \cite{shirazi2021directed} groups similar instances and focuses substitutions on clusters with historically high evasion rates, thereby speeding up the search. 
 \item \textit{Logical and mathematical transformations} apply algebraic operations to features drawn from benign distributions, malicious distributions, or both. One such example is Flow-Merge \cite{abusnaina2019examining} which builds adversarial examples by merging benign and malicious network flows via count accumulation and ratio-based averaging to preserve functional characteristics. Meanwhile, FIGA \cite{gressel2021feature} perturbs the top-k features (selected by information gain or Gini impurity) in proportion to the sample’s total feature sum and in directions dictated by the difference between class-wise feature means. MS-FIGA \cite{karumanchi_minimum_2023} further reduces model queries by clustering samples and reusing cluster representatives for perturbation. In addition, simpler attacks \cite{agarwal2021black, yuan2024multi} manipulate binary features through bitwise OR operations or flips to benign categories, and apply fixed constant or logarithmic-exponential transforms on continuous features to reduce detection scores. Finally, \citet{debicha2023adv} performs gradient estimation by using the sign of the difference between benign feature means and the original malicious instance to guide perturbations. 
 
 \item \textit{Statistical pattern-based methods}, such as A2PM \cite{vitorino_adaptative_2022}, maintain dynamic interval statistics for continuous features and store valid categorical combinations. They apply additive or subtractive modifications within moving-average intervals or swap coherent categorical sets to preserve semantic consistency.
\end{itemize}

\textbf{Meta-heuristic attacks}, on the other hand, treat adversarial example generation as a multi-objective optimization with one or more of these objectives, such as evasion success, perturbation magnitude, and feasibility of adversarial examples in the real world. Evolutionary algorithms and swarm-based strategies have been applied to various domains. For example, CoEVa2 \cite{ghamizi2020search} formulates credit-scoring evasion as a four-objective genetic algorithm that balances misclassification, minimal perturbation, domain constraints, and overdraft gains. Although it requires manual weight tuning for each objective. MOEVA \cite{simonetto2021unified} abstracts this framework to any tabular task by automatically balancing objectives without domain expertise. Similarly, \citet{alhajjar2021adversarial} compares genetic algorithms and particle swarm optimization for network intrusion detection, using a voting ensemble as the fitness function but considering only misclassification and overlooking domain constraints. 

\textbf{Other approaches} include a diverse range of techniques, among them constrained optimization with algorithms  like COBYLA that iteratively perturb original examples under linear and non-linear constraints. OptAttack \cite{chen2020generating} and AdversSpam \cite{concone2024adverspam} both use COBYLA to satisfy complex constraints ranging from semantic dependencies to statistical correlations, while aiming for evasion. However, these approaches come with the cost of scaling poorly with input dimensionality. On a different approach \citet{kireev_adversarial_2023} constructs a graph whose nodes are feasible states of the input data and edges correspond to allowable feature changes weighted by their real-world modification costs. The algorithm then finds the cheapest path that flips the model’s decision. Meanwhile, MISLEAD \cite{khazanchi2024mislead} uses SHAP values to select the direction of perturbation and a binary search to compute the minimal perturbation needed for misclassification. Finally, researchers have extended gradient-free image-based attacks to tabular data, similarly to what we saw with gradient-based attacks. \citet{cartella_adversarial_2021} modify Zoo, Boundary, and HopSkipJump algorithms with custom classification-error thresholds, perturbation norms reflecting human inspection probabilities, and strict type checks. Similarly, \citet{alhussien2024constraining} refines Zoo for the network intrusion domain, to handle structural and inter-feature constraints.

\begin{table}[t!]
\centering
\scriptsize
\caption{Overview of Gradient-free Adversarial Attacks on Tabular Data.}
\label{tab:gradient_free_attacks}
\footnotesize
\begin{tabular}{@{}p{0.5cm}p{0.5cm}p{1.8cm}p{1.5cm}p{1cm}p{2cm}p{2.5cm}p{1.5cm}p{0.8cm}@{}}
\toprule
\textbf{Ref.} & \textbf{Year} & \textbf{Attack} & \textbf{Task} & \textbf{Target} & \textbf{Knowledge} & \textbf{Description} & \textbf{Domain}& \textbf{Code} \\ \midrule
\cite{shirazi2019adversarial} & 2019 & -- &
Classification& Yes & Gray-box & Combinatorial & Cybersecurity  & No \\
\cite{abusnaina2019examining} & 2019 & Flow-Merge &
Classification& Yes \& No  & Black-box (query) & Merge features & Cybersecurity  & No \\
\cite{chen2020generating} & 2020 & OptAttack &
Classification & No & Black-box (transfer) & Cons. Optimization & Cybersecurity & No \\
\cite{ghamizi2020search} & 2020 & CoEvA2 &
Classification& No & Black-box (query) & Genetic Algorithm & Finance & Yes \\
\cite{alhajjar2021adversarial} & 2020 & -- &
Classification& Yes & Black-box (query) & Genetic Algorithm & Cybersecurity & No \\
\cite{alhajjar2021adversarial} & 2020 & -- &
Classification& Yes & Black-box (query) & Particle Swarm Opt. & Cybersecurity & No \\
\cite{shirazi2021directed} & 2021 & -- &
Classification& Yes & Gray-box & Combinatorial & Cybersecurity  & No \\
\cite{gressel2021feature} & 2021 & FIGA &
Classification& Yes & Gray-box & Rule based  & General  & No \\
\cite{cartella_adversarial_2021} & 2021 & -- &
Classification & No & Black-box (query) & Modified Zoo  & Finance  & No \\
\cite{cartella_adversarial_2021} & 2021 & -- &
Classification& No & Black-box (query) & Modified HopSkipJump  & Finance  & No \\
\cite{cartella_adversarial_2021} & 2021 & -- &
Classification & No & Black-box (query) & Modified Boundary  & Finance  & No \\
\cite{simonetto2021unified} & 2021 & MOEVA &
Classification& No & Black-box (query) & Genetic Algorithm  & General & Yes \\
\cite{agarwal2021black} & 2021 & -- &
Classification& No & Black-box (query) & XOR and constant $\delta$ & Finance & No \\
\cite{vitorino_adaptative_2022} & 2022 & A2PM &
Classification& Yes \& No  & Black-box (query) & Statistical pattern & Cybersecurity & Yes \\
\cite{yuan2024multi} & 2022 & Multi-SpacePhish &
Classification& Yes & Black-box (query) & Flip to benign cat. & Cybersecurity & Yes \\
\cite{kireev_adversarial_2023} & 2022 & -- &
Classification& No & Black-box (transfer) & Graphical framework & General & No \\
\cite{karumanchi_minimum_2023} & 2023 & MS-FIGA &
Classification& Yes & Gray-box & Rule based  & General  & No \\
\textcolor{blue}{\cite{debicha2023adv}} & 2023 & -- &
Classification& No & Gray-box & Gradient estimation & Cybersecurity & No \\
\cite{concone2024adverspam} & 2024 & -- &
Classification & Yes & Gray-box & Cons. optimization & Cybersecurity & Yes \\
\cite{khazanchi2024mislead} & 2024 & -- &
Classification & Yes \& No & Black-box (query) & Binary search & Cybersecurity & No \\
\cite{alhussien2024constraining} & 2024 & -- &
Classification & Yes \& No & Black-box (query) & Constrained Zoo & Cybersecurity & Yes \\
\bottomrule
\end{tabular}
\end{table}

%%%%%%%%%%%%%%%%%%%%%%%%%%%%%%%%%%%%%%%%%%%%%%%%%%%
%%%%%%%%%%%%%%%%%%%%%%%%%%%%%%%%%%%%%%%%%%%%%%%%%%%%%%%%%%%%%%%%%%%%%%%%%%%%%%%%%%%%%%%%%%%%%%%%%%%%%%
Gradient-free attacks on tabular data are characterized more by diversity than by shared design principles. Rule-based methods are mostly manually crafted, using simple, domain-specific heuristics like bit-flipping or count merging. While efficient, these attacks rarely generalize, and there’s little effort to formalize their logic. Meta-heuristic methods offer broader applicability, but adaptations of algorithms like genetic or swarm search remain limited. A few gradient-free attacks from the image domain have been adapted to tabular settings with added feasibility constraints, but far fewer than in gradient-based attacks, where such adaptations are the norm. Despite some progress, the field still lacks a clear understanding of what makes a gradient-free attack effective under realistic tabular constraints.

\subsubsection{Learning-based approaches}

Table \ref{tab:learning_attacks} summarizes learning-based attacks on tabular data. Notably, GAN-based methods dominate this category, with only two approaches deviating from this trend with alternative architectures.

\textbf{GAN-based approaches.} These were among the earliest methods developed for adversarial attacks on tabular ML. Among the reviewed works, \citet{lin2022idsgan} is both the earliest and the most widely cited. 
The studies under our review that use GANs generally aim to perturb non-functional features while preserving the underlying data semantics. While the goal is consistent across studies, architectural and input design choices vary. For example, (C-)WGANs are employed by \cite{lin2022idsgan, zhao2021attackgan, duy2023investigating, sun2023gpmt, bai2024adversarial}, CTGAN by \citet{asimopoulos_breaching_2023}, whereas others like \cite{usama2019generative, chen2020generating} rely on vanilla GANs. Input strategies to the generator also differ: some use original instances \cite{usama2019generative, chen2020generating, sun2023gpmt}, \citet{alhajjar2021adversarial} adds random noise to mutable sub-vectors, \cite{zhao2021attackgan, asimopoulos_breaching_2023} uses pure noise, and others combine noise with selected features (e.g non-functional ones) \cite{lin2022idsgan, duy2023investigating, wang2024ids, bai2024adversarial}.  Exceptionally,\citet{parfenov_investigation_2023} takes a non-traditional approach to using GANs as adversarial generators. Their method involves inverting the values of the target binary feature and providing deliberately false class labels during their CGAN training, leading to the generation of synthetic adversarial examples that closely resemble real data while misrepresenting true classifications. Finally, \citet{dyrmishi2024deep} adapts tabular data generators like CTGAN, TableGAN, and TVAE for adversarial use, incorporating a constraint-repair layer to ensure generated samples adhere to domain-specific linear constraints. 

\textbf{Others.} Only two studies \cite{duan_attacking_2024,de2021fixed} do not use GANs as their learning model. \citet{duan_attacking_2024} propose a multi-step attack that begins by selecting, for each target sample, the top-k most similar samples from the opposite class using KNN applied to their embedded feature representations. Following that, an encoder-decoder network generates perturbations, guided by classification loss to flip the prediction and similarity loss to preserve realistic feature distributions. The final adversarial example combines adjusted continuous features and discrete ones selected from the candidate set using KNN. Meanwhile, \citet{de2021fixed} leverages reinforcement learning (RL) to generate adversarial perturbations. The RL agent receives an input instance and outputs a perturbation based on its policy. It gets a reward of 1 if the classifier is fooled, and 0 otherwise. Each iteration consists of a single step, and the agent continually refines its policy to improve adversarial generation.
\begin{table}[t!]
\centering
\caption{Overview of Learning-based Adversarial Attacks on Tabular Data.}
\label{tab:learning_attacks}
\footnotesize
\begin{tabular}{@{}p{0.5cm}p{0.5cm}p{1cm}p{1.5cm}p{1cm}p{2cm}p{1.8cm}p{1.5cm}p{0.8cm}@{}}
\toprule
\textbf{Ref.} & \textbf{Year} & \textbf{Attack} & \textbf{Task} & \textbf{Targeted} & \textbf{Knowledge} & \textbf{Description} & Domain & \textbf{Code} \\ 
\midrule
\cite{lin2022idsgan}  & 2018 & IDSGAN & Classification & No & Black-box (query) & Adversarial GAN & Cybersecurity & No \\
\cite{usama2019generative} & 2019 & -- & Classification & No & Black-box (transfer) & Adversarial GAN & Cybersecurity & No \\
\cite{chen2020generating} & 2020 & -- & Classification & No & Black-box (transfer) & Adversarial GAN & Cybersecurity & No \\
\cite{zhao2021attackgan} & 2021 & attackGAN & Classification & No & Black-box (query) & Adversarial GAN & Cybersecurity & No \\
\cite{de2021fixed} & 2021 & FAST & Classification & No & Black-box (query) & Adversarial RL & Cybersecurity & No \\
\cite{alhajjar2021adversarial} & 2021 & -- & Classification & Yes & Black-box (query) & Adversarial GAN & Cybersecurity & No \\
\cite{parfenov_investigation_2023} & 2023 & -- & Classification & No & Black-box (query) & Adversarial GAN & Other & No \\
\cite{sun2023gpmt} & 2023 & GPMT & Classification & No & Black-box (transfer) & Adversarial GAN & Cyber & Yes \\
\cite{duy2023investigating} & 2023 & Fool-IDS & Classification & No & Black-box (query) & Adversarial GAN & Cybersecurity & No \\
\cite{asimopoulos_breaching_2023} & 2023 & -- & Classification & No & White-box & Adversarial GAN & Cybersecurity & No \\
\cite{duan_attacking_2024} & 2024 & CTRAttack & Classification & Yes & Black-box (transfer) & Encoder-Decoder & Other & No \\
\cite{dyrmishi2024deep} & 2024 & -- & Classification & No & Black-box (query) & Adversarial GAN & General & Yes \\
\cite{wang2024ids} & 2024 & IDS-GAN & Classification & Yes \& No & Black-box (query) & Adversarial GAN & Cybersecurity & No \\
\cite{bai2024adversarial} & 2024 & -- & Classification & Yes \& No & Black-box (transfer) & Adversarial GAN & Other & No \\
\bottomrule
\end{tabular}
\end{table}

Learning-based adversarial attacks on tabular data remain underdeveloped, despite their early promise. The field is heavily concentrated around GAN-based methods (12 out of 14), most of which target cybersecurity applications. However, these approaches tend to replicate the same high-level design with only minor architectural variations, such as different noise injection strategies or GAN variants. These differences are rarely evaluated systematically, and there is no standardized framework to compare components or isolate their contributions. Only one recent work \cite{dyrmishi2024deep} attempts a direct comparison across multiple generative architectures, revealing that most GAN-based methods still fall short of state-of-the-art gradient or search-based attacks. Additionally, many of these learning-based attacks do not reference or compare to one another, nor do they clearly build upon the broader lineage of learning-based attacks in the image domain. This lack of cumulative progress and critical self-positioning has led to a fragmented literature with marginal gains and no comparative benchmarking. 

%%%%%%%%%%%%%%%%%%%%%%%%%%%%%%%%%%%%%%%%%%%%%%%%%%%
%%%%%%%%%%%%%%%%%%%%%%%%%%%%%%%%%%%%%%%%%%%%%%%%%%%%%%%%%%%%%%%%%%%%%%%%%%%%%%%%%%%%%%%%%%%%%%%%%%%%%%

\subsubsection{Hybrid approaches}

\begin{table}[t!]
\centering
\scriptsize
\caption{Overview of Hybrid Adversarial Attacks on Tabular Data.}
\label{tab:hybrid_attacks}
\footnotesize
\begin{tabular}{@{}p{0.5cm}p{0.5cm}p{1cm}p{1.5cm}p{1cm}p{1.8cm}p{3.3cm}p{1cm}p{0.8cm}@{}}
\toprule
\textbf{Ref.} & \textbf{Year} & \textbf{Attack} & \textbf{Goal} & \textbf{Targeted} & \textbf{Knowledge} & \textbf{Description} & Domain & \textbf{Code} \\ \midrule
\cite{wang2020attackability} & 2020 & OPMGS & Classification & No & White-box & Gradient + Greedy Search & General & Yes \\
\cite{kumar_evolutionary_2021} & 2021 & ESPA & Classification & No & Black-box (query) & Encoders + GA & Finance & No \\
\cite{malik_tetraa_2023} & 2023 & TETRAA & Classification & No & Black-box (query) & Encoders + GA & General & No \\
\cite{bao2023towards} & 2022 & FEAT & Classification & No & White-box & OMP + MAB & General & Yes \\
\cite{pandey_improving_2023} & 2023 & - & Classification & No & Black-box (query) & GAN + Boundary/HopSkipJump & Finance & Yes \\
\cite{simonetto2024constrained} & 2023 & CAA & Classification & No & White-box & CAPGD + MOEVA & General & Yes \\
\bottomrule
\end{tabular}
\end{table}

Some works propose hybrid adversarial attack methods that integrate two or more approaches: gradient-based, gradient-free, and learning-based. An overview of these methods is provided in Table~\ref{tab:hybrid_attacks}.

For example, \citet{kumar_evolutionary_2021}  evolved a population of encoders that take a benign input and generate specialized perturbations tailored to that input. The evolution process, guided by a genetic algorithm, optimizes these encoders over multiple generations, ultimately preserving the best-performing encoder to generate adversarial examples. This approach was later enhanced by \citet{malik_tetraa_2023} and transformed into a general method for tabular data rather than being limited to payment systems. The improved version supports categorical features and incorporates a binary search-based method, further enhancing its effectiveness. Generative models have also been combined with traditional attack techniques. \citet{pandey_improving_2023} proposed a GAN-based Feature Selector Network to identify a minimal set of vulnerable features most susceptible to adversarial attacks. The selected minimal feature set is then used to craft adversarial examples using traditional attacks like Boundary \cite{brendel2017decision} and HopSkipJump \cite{chen2020hopskipjumpattack}. 

\citet{wang2020attackability} presents  OMPGS attack for discrete data. Categorical features are first mapped into continuous embedding vectors, allowing the model to compute gradients. At each step, the algorithm selects the feature with the highest gradient magnitude to maximize the impact on the attack outcome. Then the value is modified through a greedy-search strategy until a misclassification is achieved, or a stop condition is met.  \citet{bao2023towards} later extended this with FEAT, which uses orthogonal matching pursuit to rank feature influence and a multi-armed bandit approach with an Upper Confidence Bound (UCB) strategy to guide the search for the most effective adversarial modifications. 

Another interesting hybrid strategy named Constrained Adaptive Attack (CAA) was proposed by \cite{simonetto2024constrained}, building upon their gradient-based attack CAPGD \cite{simonetto2024towards}  and their evolutionary attack MoEvA \cite{simonetto2021unified}. The core idea of CAA is to leverage the efficiency of gradient-based attacks and the success rate of search-based attacks. The CAA approach first applies CAPGD to generate valid adversarial examples at a low computational cost. If CAPGD fails on an original example, CAA executes the slower but more effective genetic algorithm MOEVA method.

Hybrid attacks aim to combine the speed and efficiency of gradient-based methods with the robustness of search-based techniques or the semantic modeling capabilities of generative approaches. In doing so, they enhance attack diversity, helping to bypass defenses specifically tailored to one attack method. These methods often use one method to guide or initialize another, such as running gradient attacks first and switching to search when they fail, or using generative models to identify key features before applying classical attacks.  While this makes them more flexible, they lack on general frameworks or benchmarking. There is limited evidence that these combinations consistently outperform strong single-method attacks. Still, hybrid strategies point toward a useful direction: matching attack components to the structure of the data and the task, rather than relying on one-size-fits-all methods.

\begin{tcolorbox}[
colframe=black,          % Dark border for contrast
colback=gray!5,         % Light gray background (soft & elegant)
coltitle=white,         % White title text
colbacktitle=purple!50!black, % Deep purple title background
title=\bfseries \Large Attack methodologies, center title, % Bold title text
sharp corners=south,    % Sharp bottom corners
fonttitle=\bfseries     % Bold title
]

Across attack families, tabular adversarial research reveals distinct patterns and challenges.  \textit{Gradient-based methods} mostly adapt image attacks with added feasibility constraints. \textit{Gradient-free approaches} show more diversity but lack shared principles, limiting broader applicability. \textit{Learning-based attacks}, dominated by GANs in cybersecurity, struggle with minimal innovation and fragmentation without standardized evaluation frameworks. \textit{Hybrid attacks} offer flexibility by combining strengths of multiple methods, but need more rigorous validation to prove consistent gains. nevertheless, domain-specific constraints remain a challenge across all optimization strategies, and open-source support is limited, with only 21 out of 61 attacks having public code.
\end{tcolorbox}

\subsection{RQ3: Practical considerations for adversarial attacks on tabular data}

\label{sec:RQ3}

To evaluate how existing tabular‐data attack studies address the practical requirements of real‐world deployment  outlined in Section \ref{subsec:aspects}, we reviewed each paper against eight key dimensions: \emph{efficacy}, \emph{efficiency}, \emph{transferability}, \emph{feasibility}, \emph{semantic preservation}, \emph{plausibility}, \emph{defense awareness} and \emph{dataset suitability}. For every dimension in each paper, we assigned one of three labels:

\begin{itemize}[noitemsep, topsep=0pt]
  \item \textbf{Considered.} The study explicitly integrated this aspect into its attack design or empirical evaluation. For example, by measuring runtime (efficiency) or verifying the stealthiness of adversarial examples (plausibility).
  \item \textbf{Acknowledged.} The dimension is not directly addressed in the design or experiments, but the authors explicitly recognize its importance. This recognition could appear in sections like “Limitations” and “Future Work” or as brief but explicit commentary in the main text. 
  \item \textbf{Not considered.} The paper makes no explicit mention of the dimension, neither incorporating it into the work nor discussing its relevance.
\end{itemize}

This categorization measures
how often practical considerations are addressed in the reviewed literature and how often authors acknowledge the remaining gaps.  
We provide a detailed mapping of what constitutes Considered, Acknowledged, and Not considered for each practical dimension in Table \ref{tab:rq3_marking_explained} (Appendix \ref{app:rq3}). After applying these labels to each paper in our survey, we compiled a comprehensive mapping of the results presented in Table~\ref{tab:rq3_detailed} (Appendix~\ref{app:rq3}). A high-level summary is illustrated in Figure~\ref{fig:aspects_overview}, which serves as a starting point for the in-depth analysis that follows.

\begin{figure}[ht!]
    \centering        \includegraphics[width=0.8\linewidth]{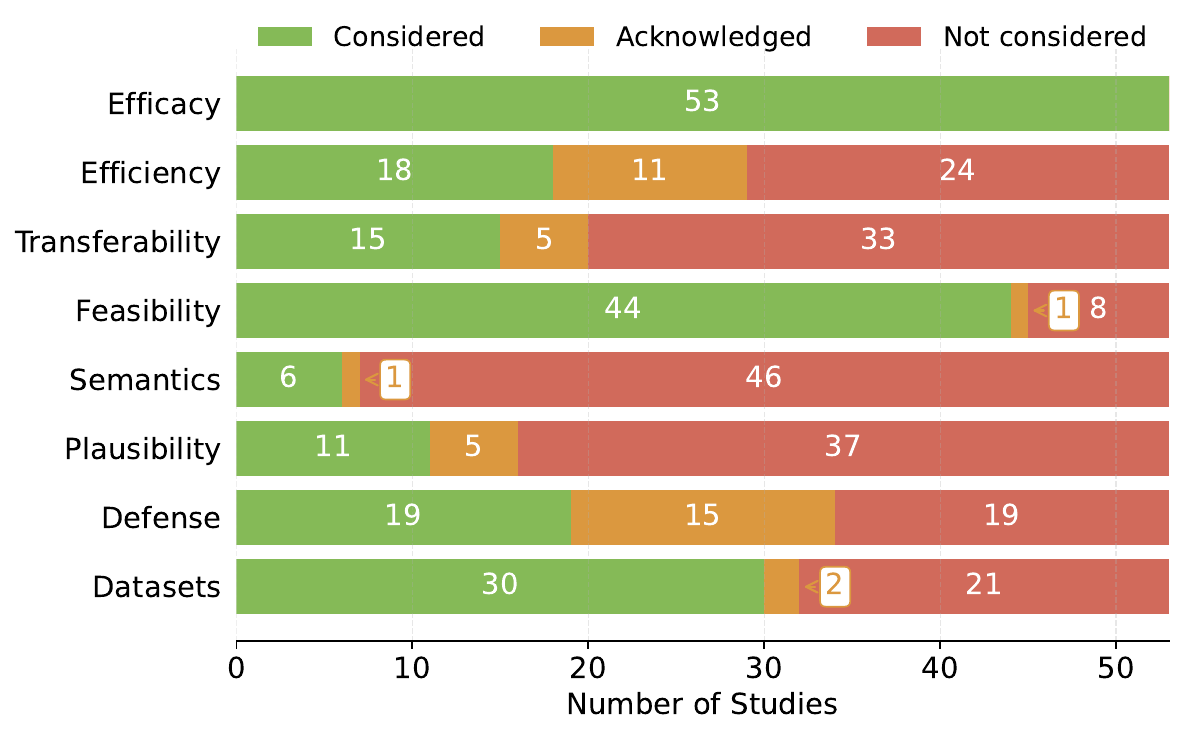}
        \caption{Aggregated statistics regarding practical considerations addressed across studies.}
        \label{fig:aspects_overview}
\end{figure}

The results show that all 53 surveyed studies addressed the efficacy criteria. We hypothesize that, in order to be considered for publication, any new attack must demonstrate the ability to successfully bypass tabular ML models while achieving a competitive success rate. As a result, efficacy has effectively become the de facto standard for publication in this field. However, this strong emphasis on efficacy often comes at the expense of other practical considerations that affect real-world applicability.  For example, while 44 studies do address the feasibility of generating adversarial examples, far fewer account for their semantic integrity (6 studies) or plausibility (11 studies). This oversight increases the risk that such examples would be easily detected by simple data validation techniques or manual review. 
Transferability is similarly underexplored, only 15 studies test whether adversarial examples remain effective across different models. Furthermore, just 19 studies evaluate performance against defense mechanisms, highlighting a limited understanding of robustness in secured environments. Finally, although 30 studies consider the suitability of the datasets used for evaluation, inconsistent practices make it difficult to compare results across studies or generalize findings.

This aggregated overview already provides relevant insights in the priorities given to each criteria in the current literature. The fact that efficacy is always covered hints that this is a minimal requirement of the community. The relatively high attention to feasibility likely reflects two factors: first, the explicit requirements set by leading security venues such as S\&P and Usenix, which promote this line of research; second, the inherent challenges of tabular settings that necessitate tailored attacks accounting for their unique feasibility constraints. In contrast, the limited consideration of more challenging criteria, such as transferability, plausibility, and semantics, may hint at the lack of awareness of the community about the importance of these criteria.  Meanwhile, the partial coverage of efficiency and defense awareness criteria suggests that while the community recognizes their relevance, these aspects are rarely addressed adequately. This could be  caused by unsuitable baselines and benchmarks, or the need for expensive/complex approaches. 

Lastly, we analyze  the number of practical aspects addressed by each of the surveyed studies in Figure \ref{fig:dimensions_distribution}. Most studies consider around four dimensions, with the majority falling between two and six. This suggests that while researchers acknowledge multiple evaluation criteria, it is uncommon for studies to comprehensively cover all practical considerations. Notably, only one study \cite{asimopoulos_breaching_2023} focuses exclusively on efficacy, highlighting a narrow approach. In contrast, two studies \cite{debicha2023adv, duan_attacking_2024} stand out by addressing seven aspects, demonstrating a more well-rounded  approach. This variability highlights the absence of standardized practices and suggests room for more thorough and balanced attack design and evaluation in future work.

\begin{figure}[ht!]
    \centering        \includegraphics[width=0.7\linewidth]{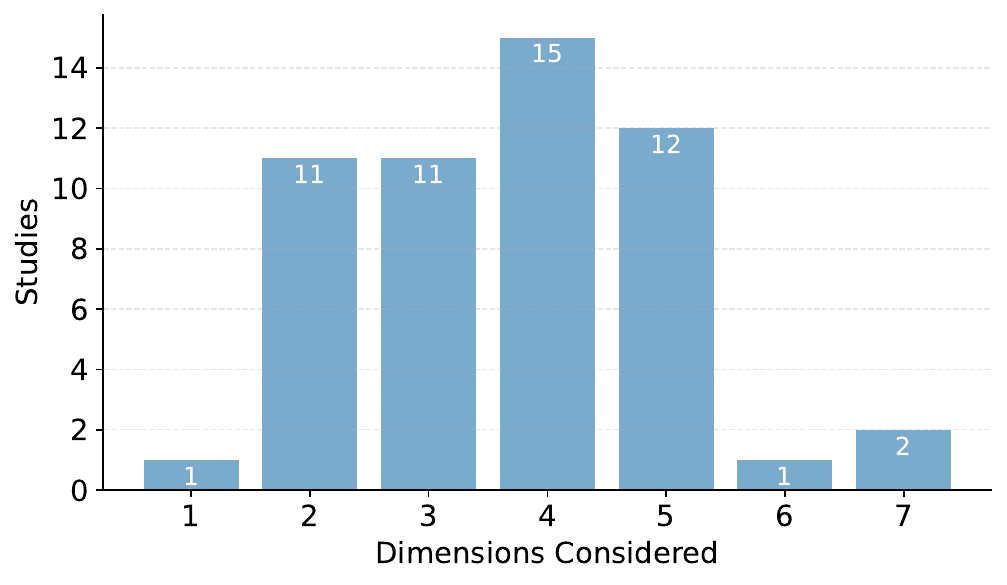}
        \caption{Distribution of Considered Practical Dimensions Across Studies.}
        \label{fig:dimensions_distribution}
\end{figure}

We explore in the following subsections our hypotheses through a detailed analysis of each practical consideration, and highlight a few representative works for each criterion.

\subsubsection{Efficacy.}
Figure \ref{fig:aspects_overview} shows that all reviewed studies explicitly consider attack efficacy, which is a central objective in the design of attack strategies, as higher success rates are inherently preferred. Most studies report efficacy using standard metrics related to success rate (ASR) or robust accuracy (RA). While accuracy is more commonly reported, four studies \cite{usama2019generative, alhajjar2021adversarial, debicha2023adv,asimopoulos_breaching_2023} also incorporate additional metrics like the F1 score and the ROC curve under both benign and adversarial conditions. This is particularly important for unbalanced datasets, where accuracy alone may be insufficient. In addition to these, six studies \cite{lin2022idsgan, zhao2021attackgan, sun2023gpmt, duy2023investigating, wang2024ids, bai2024adversarial} report Evasion Increase Rate (EIR), calculated as $\text{EIR} = \frac{\text{DR}_{\text{adv}}}{\text{DR}_{\text{att}}}$, where $\text{DR}_{\text{adv}}$ is the model's accuracy on adversarial samples and $\text{DR}_{\text{att}}$ is the accuracy on original samples. Furthermore, three studies \cite{kumar_evolutionary_2021, pandey_improving_2023, malik_tetraa_2023} report the Mean Probabilities of Successful Attacks (MPSA), where higher values indicate more effective attacks.

In the context of regression-based adversarial robustness evaluation,  \citet{kong2023adversarial} adopts the classical form of Mean Absolute Percentage Error (MAPE), where the error is computed as the deviation between the original and adversarial predictions. Similarly, \citet{gupta_adversarial_2021} uses Symmetric MAPE (SMAPE), an updated version of MAPE that addresses its sensitivity to small denominators. In addition to SMAPE, they report Fooling Error, defined as the average difference between the model's original and adversarial predictions, and Mean Absolute Error (MAE), which measures the average deviation between the adversarial predictions and the true target values.

Overall, most efficacy evaluation criteria focus on model error rates and do not take into account whether the model achieves its real-world objectives. Notably, \cite{ghamizi2020search} is an exception, as it measured success by the amount of money lost by the bank. However, such real-world impact assessments remain uncommon, and there is still little research connecting standard machine learning performance metrics to meaningful practical outcomes.

\subsubsection{Efficiency.} Attack efficiency is addressed in 18 studies, either as a primary focus or as part of the evaluation framework. An additional 11 studies acknowledge its importance or propose it as a direction for future work, while 24 studies do not mention it at all. Four works \cite{wang2020attackability, vitorino_adaptative_2022, malik_tetraa_2023, karumanchi_minimum_2023} explicitly aim to develop more efficient adversarial attack methods. Among them, \citet{karumanchi_minimum_2023} explores the minimum number of training samples required to craft a successful attack, focusing more on the dimension of data efficiency.

The evaluation of attack efficiency varies significantly between studies. In most cases, efficiency is assessed indirectly through average runtime. For example, \citet{vitorino_adaptative_2022} reports the average time per iteration of their A2PM attack. \citet{duan_attacking_2024} compares the efficacy of their method against state-of-the-art attacks under a strict one-hour time constraint. \citet{kireev_adversarial_2023} introduces a novel efficiency metric based on the success-to-time ratio. Similarly, \citet{sheatsley2021robustness} assesses the scalability of their constraint-learning pre-attack phase by measuring its runtime on increasingly large datasets. Other studies, such as \citet{kumar_evolutionary_2021} and \citet{malik_tetraa_2023},  rely on the number of queries as the primary measure of efficiency. This is especially a relevant metric in black-box settings, where query limits are often enforced. In such cases, models may also include detection mechanisms that monitor and block suspicious query patterns indicative of adversarial behavior \cite{chen2020stateful}. Finally, two other studies by \citet{pandey_improving_2023} and \citet{bao2023towards}, report both query counts and runtime for a more comprehensive view of attack efficiency.

Overall, there is no standard metric on measuring the efficiency of adversarial attacks and little interpretation on what these metric means for a real-world context. While runtime is the most commonly reported metric, query count remains critical in black-box contexts with detection-aware defenses. The diversity in how efficiency is measured suggests a need for standardized benchmarking criteria, as inconsistent reporting hinders comparison across studies.

\subsubsection{Transferability.} Despite its importance in black-box attack scenarios, transferability of adversarial examples is directly evaluated in only 15 studies, with an additional 5 studies acknowledging its significance, and 32 studies not considering it at all. Among the 15 studies that do evaluate transferability \cite{sheatsley2020adversarial, tian2020exploring, alhajjar2021adversarial, chen2020generating, cartella_adversarial_2021, shirazi2021directed,  ju_robust_2022, mathov_not_2022, chernikova_fence_2022, simonetto2024constrained,  debicha2023adv, duan_attacking_2024, concone2024adverspam, alhussien2024constraining, bai2024adversarial}, all report the transferability rate $TR$ as part of their evaluation. Meanwhile, \citet{alhussien2024constraining} explicitly identifies the study of transferability as a primary objective of their work.

The study of adversarial example transferability is well-established in computer vision, with numerous approaches proposed in the literature \cite{gu2023survey, djilani2024robustblack}. However, this line of research gained significant attention only after substantial progress had been made in understanding and mastering the white-box setting. We hypothesize that a similar trend is occurring in the context of tabular machine learning. Our findings suggest that the field is still primarily focused on foundational aspects such as attack efficacy and effectiveness, which may explain why transferability remains underexplored. Nonetheless, the techniques developed in computer vision offer promising directions for adaptation to tabular data.

\subsubsection{Feasibility.} 
It is  addressed in 44 of the 53 surveyed studies, ignored in 8 studies, and explicitly mentioned as an area for future work by \citet{wang2020attackability}. It remains unclear whether these omissions come from the simplicity of the datasets and domains under study, or from the challenges associated with enforcing such constraints. Nevertheless, the absence of feasibility considerations often leads to unrealistic adversarial examples, particularly in complex domains, thereby limiting the practical relevance of these approaches.

Among the studies that address feasibility, the level of consideration varies significantly from simplistic assumptions to detailed modeling of real-world constraints. To reflect this variation, we conducted a more fine-grained analysis over three type of constraints: 
mutability constraints, structural constraints, and inter-feature relationships.\footnote{It is noteworthy that when feasibility is mentioned only as a limitation or a direction for future work, it is often unclear which specific type of constraint is being referred to. Therefore, in our classification, we conservatively categorize such cases as “not considered” for each specific constraint type.} Figure~\ref{fig:feasibility} offers an overview of how frequently these constraints are addressed in the literature.

\begin{figure}[ht!]
    \centering
    \begin{minipage}{0.49\textwidth}
        \centering
        \includegraphics[width=\linewidth]{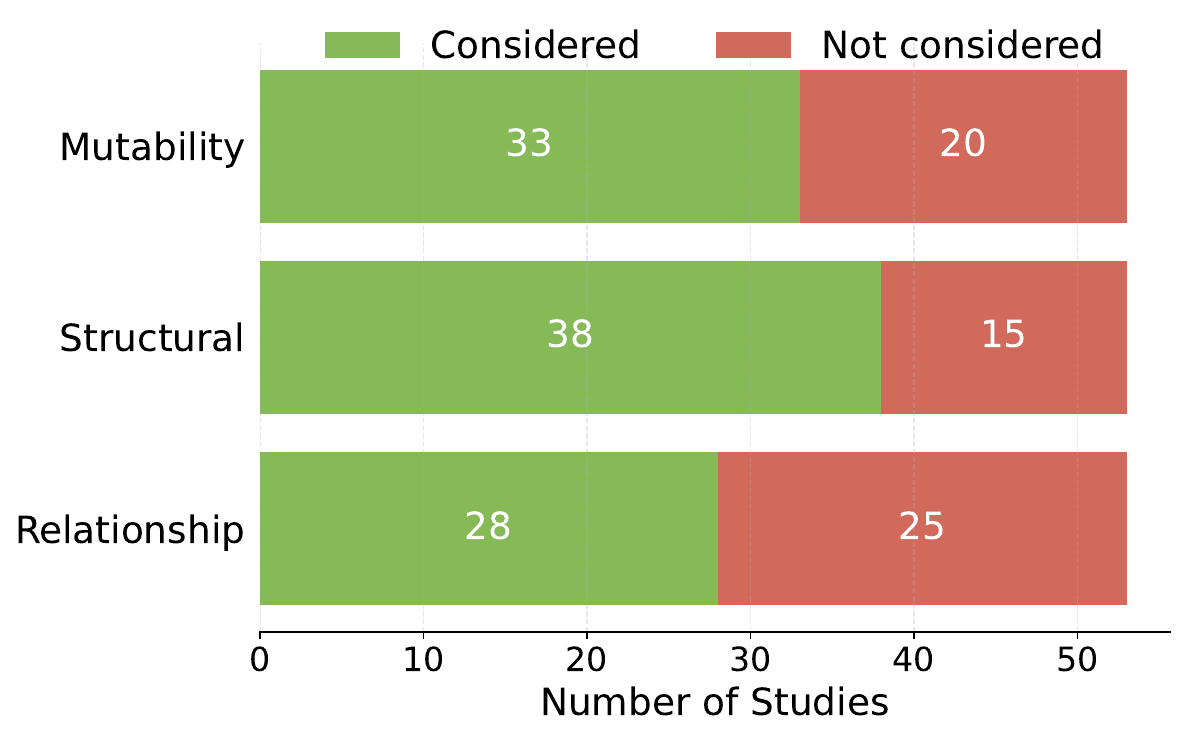}
   \caption{Overview of addressed feasibility constraints. }
    \label{fig:feasibility}
    \end{minipage}
    \hfill
    \begin{minipage}{0.49\textwidth}
        \centering
        \includegraphics[width=0.6\linewidth]{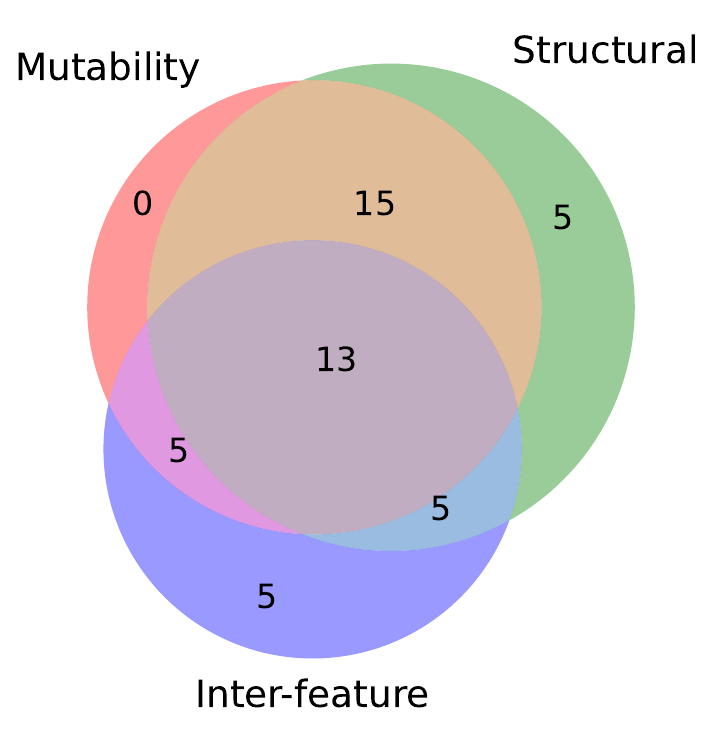}
        \caption{Overlap between addressed feasibility constraints.}
        \label{fig:feasibility_venn}
    \end{minipage}
\end{figure}

We provide below a more detailed analysis of each category.

\begin{itemize}
 \item \textit{Mutability constraints} were considered in 33 out of 53 studies. These primarily prevent perturbation of immutable or user-inaccessible features, such as age or blood type. Notably, five studies \cite{usama2019generative, alhajjar2021adversarial, sun2023gpmt, duy2023investigating, wang2024ids} restrict perturbations to features known to have low functional impact to avoid unrealistic manipulations. While this strategy helps maintain feasibility, it also reduces the attack search space and may weaken the attack's effectiveness.
\item \textit{Structural constraints}  were the most commonly applied, appearing in 38 studies. These include enforcing valid feature types, allowable ranges (e.g., min-max bounds), and encoding formats (e.g., ensuring categorical features remain valid). Due to their relative simplicity and generalizability, these constraints often form a baseline feasibility condition in tabular attack methods.
\item \textit{Inter-feature relationships} were addressed in 28 studies, making them the least frequently enforced, likely due to their complexity. They are enforced via two primary strategies:

    \begin{itemize}
        \item  \textit{Explicit enforcement.} This approach is used in 19 studies and includes methods such as projection-based techniques (14 studies), constraint solvers like SAT and DPLL \cite{sheatsley2021robustness, ben-tov_cafa_2024}, and differentiable penalty terms \cite{simonetto2021unified, simonetto2024constrained, simonetto2024towards}. When the sourced constraints are sound and complete, explicit enforcement provides strong guarantees of validity. However, if constraints are incomplete or poorly specified, this can lead to invalid adversarial examples or overly restrict the diversity of the attack space. In terms of how these constraints are sourced, 14 studies manually defined them using domain-specific knowledge, such as logical rules. Five studies \cite{sheatsley2021robustness,ben-tov_cafa_2024, tian2020exploring, nandy_non-uniform_2023, concone2024adverspam} relied on automated extraction techniques, including linear regression, correlation analysis, and logical rule mining approaches (e.g., Valiant, FastADC). Lastly, \citet{tian2020exploring} employed a hybrid strategy that combined manual inspection with automated extraction.
    \item \textit{Implicit enforcement.} A total of 13 studies adopt implicit enforcement strategies, most commonly through the use of generative models. These models aim to preserve the joint feature distribution of the input space, thereby implicitly promising to maintain feasibility during adversarial example generation. However, generative models for tabular data often struggle to enforce complex inter-feature dependencies. For example, \citet{stoian2024realistic} report that up to 99\% of generated samples violate feasibility constraints. To mitigate this issue, recent work by \citet{dyrmishi2024deep} proposes hybrid strategies that integrate generative approaches with post-processing steps to reintroduce hard constraints.
    \end{itemize}
\end{itemize}

Figure \ref{fig:feasibility_venn} presents a Venn diagram illustrating the number of studies that consider each feasibility constraint and their intersections. Notably, 13 studies address all three constraints, while 5 studies focus exclusively on structural constraints and another 5 only on inter-feature relationships.

Our review shows that achieving the feasibility criteria involves research across various fields of combinatorial optimization, constraint reasoning, and neuro-symbolic learning. Each attack method that enhances feasibility presents distinct strengths and trade-offs: some can cover complex scenarios but with mainly empirical validation, while others enforce domain constraints with formal guarantees, albeit within more limited contexts.

% \cite{sheatsley2021robustness,ben2024cafa, tian2020exploring, nandy2023non, concone2024adverspam}.

\subsubsection{Semantics.} Semantic preservation between original and adversarial examples was explicitly considered in only 6 out of the 53 reviewed studies. One  study \cite{kireev_adversarial_2023} acknowledged semantic consistency as an aspect, while the remaining 47 studies did not consider it. Some studies aim to preserve semantics by avoiding modifications to features strongly associated with class-specific behaviors, as seen in \cite{sun2023gpmt, debicha2023adv, chen2020generating}. Another line of work followed by \citet{ mathov_not_2022}, aim to maintain semantic correlation with the target variable by enforcing statistical measures such as Pearson correlation coefficients  in their adversarial example generation process. \citet{sheatsley2021robustness} ensures semantic preservation by introducing class-specific adversarial constraints that restrict feature modifications within observed class-specific bounds. These bounds are defined as the set of valid categorical values or the range (min–max) of continuous features for each class. \citet{teuffenbach2020subverting} argues that minimizing perturbation can relate to semantic preservation in their NIDS application, similarly to how it does in image. Meanwhile, an interesting perspective comes from \citet{kireev_adversarial_2023}. The authors acknowledge that semantic preservation is often considered a key requirement in adversarial example generation, however, they argue that this requirement is not well-suited to tabular data and even can be counterproductive. According to their reasoning, the threat is valid as long as the adversary succeeds in achieving their objective with a feasible adversarial example, even though its ground truth label might change. 

% It is worth noting that all 14 learning-based attack methods reviewed in this study are, by design, intended to preserve data distribution and maintain correlations both among features and  the target variable. However, as the respective authors did not explicitly discuss or evaluate semantic preservation, we do not categorize these studies as having considered semantic aspects. 

Semantic preservation is often under addressed, likely because authors assume their methods, especially learning-based ones, maintain semantics even without guarantees. Alternatively, they may consider it irrelevant in tabular ML or avoid the topic due to the lack of a universal definition. This is illustrated by the fact that a few studies \cite{tian2020exploring, sheatsley2020adversarial, nandy_non-uniform_2023, concone2024adverspam} refer to semantics in terms of contextual relationships within the features or correlations among them. This interpretation differs from our definition in Section \ref{subsec:aspects}, which frames semantic preservation as the retention of the original class label (e.g a malware file remains malicious after perturbation). While correlations between features contribute to the plausibility of adversarial examples, a topic we address in the following subsection, they do not necessarily ensure semantic consistency at the class level.

\subsubsection{Plausibility.}
Among the reviewed studies, 11 explicitly consider plausibility, 5 acknowledge it, and 37 omit it.  The criteria used to define plausibility vary widely across studies. 
Both \citet{ballet2019imperceptible} and \citet{cartella_adversarial_2021} assign higher perturbation budgets to features deemed less important by domain experts, and restrict changes to critical features via feature-importance weights. Meanwhile, two studies \cite{malik_tetraa_2023, duy2023investigating} use simplistic conditions for plausibility, such as rounding integer features and min-max bounds. \citet{shirazi2019adversarial, shirazi2021directed} replaces categorical features values only with feature categories that are seen in phishing instances to maintain plausibility. \citet{duan_attacking_2024} equals realism with imperceptibility and uses an encoder decoder attack method. \citet{pandey_improving_2023} justify the plausibility of adversarial examples by measuring their distance to the nearest neighbor in the training data. Lastly, \citet{concone2024adverspam} uses Lp norms as a proxy for imperceptibility.

Five studies acknowledge plausibility as an existing requirement in the literature, while challenging its strict application, especially in non-visual domains. For example, \citet{gressel2021feature} argue that plausibility should not be evaluated in the feature space, but on the adversarial example in the original data space before feature extraction. They point out that perturbations that appear significant at the feature level may remain imperceptible or irrelevant when considered from the perspective of the underlying problem. Similarly, \citet{sheatsley2020adversarial} and \citet{teuffenbach2020subverting} question the relevance of human perception in tabular domains, noting that unlike visual tasks, interpretability in structured data does not rely on human perceptual thresholds. \citet{kireev_adversarial_2023} argue that this constraint unnecessarily restricts the range of valid perturbations and the mathematical tools available for studying robustness. According to them, perceptible yet strategically chosen perturbations can still be effective. Likewise, \citet{yuan2024multi} suggest that complete "invisibility" of perturbations is not a strict requirement for their phishing detection application. This further reinforces the idea that plausibility can be context-dependent and need not tied on imperceptibility alone. 

One example that illustrates the debate around imperceptibility, plausibility, and their relationship is the use of \textit{Lp}-norms. In adversarial attacks within the image domain, $L_p-norms$ serve as a reliable proxy for assessing perturbation imperceptibility and plausibility. Following this pattern, 34 of the studies we reviewed relied on $L_p-norms$  as an attack objective, likely because they align well with existing attack frameworks—particularly those adapted from image-based attacks—and because no standardized alternatives currently exist. Additionally, 12 studies reported $L_p-norms$  statistics when evaluating the adversarial examples they generated, although with no comment on what it means for plausibility. Ultimately, plausibility, much like imperceptibility, is often implicitly assumed, deliberately avoided, or left unresolved due to a lack of consensus in the community.

% Despite their large usage, the reasoning behind their usage is rarely justified. As a result, we classify them in our "Considered" category only when authors explicitly claim to use Lp-norms for plausibility purposes, which was the case in just one out of the 34 studies.

\subsubsection{Defense awareness.} The results in Figure \ref{fig:aspects_overview} show that 19 of the reviewed studies do not evaluate their attacks against any defense mechanism. Of the 19 that do, 15 consider only a single defense, while 4 evaluate against multiple defenses. Additionally, 15 studies acknowledge defenses as an important consideration, often suggesting it as part of future work. Below, we discuss the main types of defenses evaluated in the surveyed studies.

\begin{itemize}
    \item \textit{Adversarial Training.} 
    The most commonly applied defense (18 studies) is adversarial training \cite{madry2017towards} or its variations, which remains a dominant and effective strategy even today. Adversarial training improves robustness by augmenting the training dataset with adversarial examples, allowing the model to learn from perturbed instances. 

    \item \textit{TRADES Framework.} 
    A key limitation of adversarial training is its inability to balance effectively adversarial robustness with benign accuracy. To address this, TRADES \cite{zhang2019theoretically} introduces a decomposition of the prediction error into two components: one for clean accuracy and another for regularization, which minimizes the gap between clean and adversarial representations. Two studies from \citet{nandy_non-uniform_2023} and \citet{ xu2023probabilistic} adopt TRADES to achieve a better trade-off between robustness and accuracy in tabular datasets.

    \item \textit{Adversarial detector.} This defense strategy is used by  \citet{debicha2023adv}, who designed an adversarial detector tailored for NIDS. Their method leverages an ensemble of three models, each tasked with analyzing a distinct subset of features based on their manipulability. The first model evaluates features that can be directly modified by an attacker, the second handles features that are dependent on others, and the third inspects immutable features that cannot be altered. By combining the outputs of these models, the system is able to detect and filter adversarial instances before they reach the NIDS, enhancing robustness against evasion attacks.
    
    \item \textit{Constraint Augmentation.} This  approach proposed by \citet{simonetto2021unified} enhances adversarial robustness by incorporating engineered non-convex constraints into the feature space. This method introduces new binary features derived from existing features using an XOR-based transformation. The approach ensures that original data samples inherently satisfy these constraints, whereas adversarial attacks are likely to violate them if they are not explicitly accounted for. To limit computational complexity, constraints are applied selectively to the most important mutable features, identified through Shapley value approximations. While other defenses are adapted from those used in image data, this is the only defense designed specifically for tabular data.
\end{itemize}    

Overall, while adversarial training remains the most widely used defense, alternative mechanisms inspired by the characteristics of tabular data such as constraint augmentation offer promising solutions for improving robustness. However, the limited evaluation of adversarial attacks against these defenses highlights the need for more comprehensive robustness assessments in this domain.

% Considered: Uses multiple datasets, justifies choices, discusses limitations or generalizability.

% Acknowledged: Mentions dataset limitations, biases, or potential issues, but doesn’t explore them.

% Not considered: Uses one or more datasets without discussion of representativeness, choice rationale, or limitations.

\subsubsection{Dataset Suitability.} The results revealed that 30 studies consider dataset suitability in the design of their evaluation. However, 20 papers fall into the category of not considered, either because they rely on a single dataset for evaluation or because their choice of dataset is tied to a threat model that lacks clarity. Furthermore, 2 studies explicitly acknowledge that validating their findings on datasets beyond the one they used remains an open direction for future work.

 We provide in Table~\ref{tab:datasets}, Appendix \ref{app:rq3} an overview  of 61 unique datasets used across the 53 studies under review.\footnote{These statistics are based on the datasets’ original sources (either the associated publication or official repository) to ensure consistency. Individual studies often apply unique preprocessing pipelines that may alter the number or type of features, making direct comparisons difficult without a standardized reference.} Many of these datasets come from domains where adversarial behavior is well-defined and linked to real-world concerns. Examples include intrusion detection \cite{kdd_cup_1999_data_130}, phishing classification \cite{tan2018phishing}, fraud detection \cite{ieee-fraud-detection}, and malware analysis \cite{yerima2018droidfusion}. In these cases, adversaries have clear goals, such as evading detection or manipulating predictions for financial gain. These datasets are well-suited for evaluating robustness under realistic attack scenarios. Other datasets, such as the Iris flower classification \cite{fisher1936use} and wine quality \cite {wine_109} datasets, lack a well-defined adversarial context. They are often included for methodological reasons and serve as simplified test beds for analysis or illustration. While useful for controlled experiments, they limit the relevance of robustness claims to practical applications. We hypothetize that their use is often driven by the lack of standardized benchmarks of adversarial attacks for Tabular ML. Additionally, using multiple datasets is a common practice to demonstrate the generalizability of results and is frequently encouraged during peer review.

\begin{figure}[t!]
    \centering
    \begin{minipage}{0.33\textwidth}
        \centering
        \includegraphics[width=\linewidth, height=3cm]{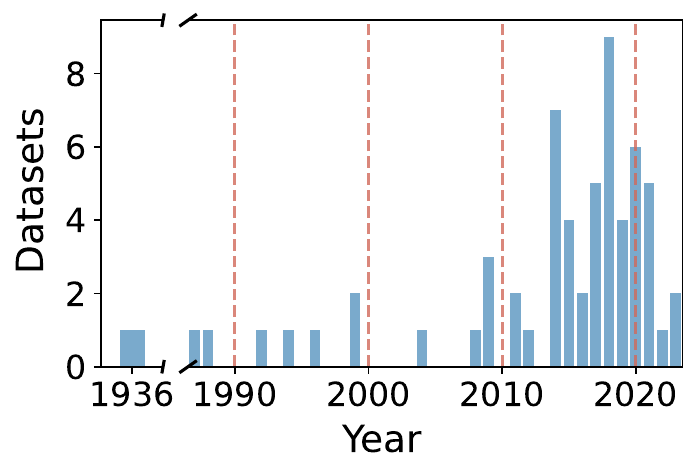}
        \caption{Distribution of datasets per year.}
    \label{fig:dataset_year_distribution}
    \end{minipage}
    \hfill
    \begin{minipage}{0.33\textwidth}
        \centering
        \includegraphics[width=\linewidth, height=3cm]{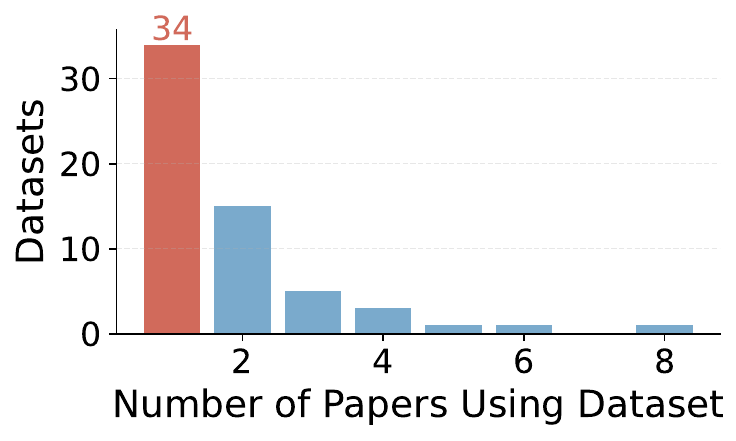}
        \caption{Dataset popularity across studies.}
        \label{fig:dataset_popularity}
    \end{minipage}
     \hfill
    \begin{minipage}{0.33\textwidth}
        \centering
        \includegraphics[width=\linewidth, height=3cm]{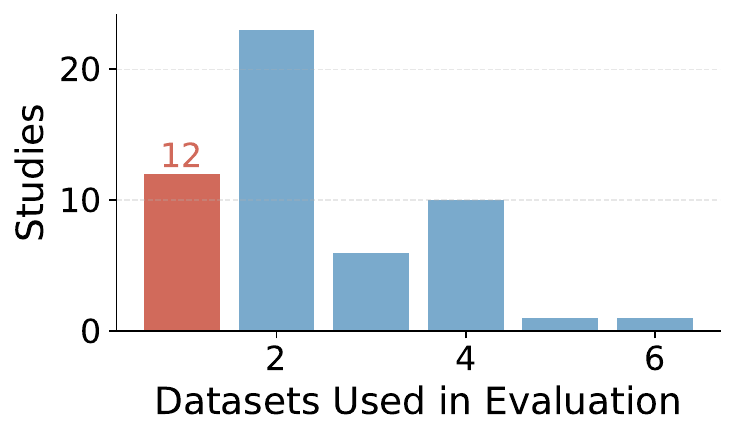}
        \caption{Datasets used per study.}
        \label{fig:eval_datasets}
    \end{minipage}
\end{figure}

The diversity of dataset usage across the literature further complicates the assessment of robustness. 32 of the 61 datasets are used by only one study (ref. Figure \ref{fig:dataset_popularity}), which indicates a fragmented evaluation landscape. While 5 datasets \cite{kdd_cup_1999_data_130, lending_club_dataset, garcia2014empirical, hannousse2021towards, ieee-fraud-detection} appear in more than four studies, there is little evidence of a consistent standard on dataset choice. This limits meaningful comparison of performance between attacks. The age of datasets is another dimension that contributes to the suitability of the dataset. Although according to Figure \ref{fig:dataset_year_distribution}, most were introduced between 2011 and 2020, there remains a continuing reliance on older datasets. Notably, 12 of them   were published before 2010, one even in 1936 \cite{fisher1936use}. Despite some of these dataset being well known (e.g. \cite{kdd_cup_1999_data_130, spambase_94}), they may no longer capture current data distributions or threat landscapes. Their ongoing use suggests a degree of inertia in dataset selection, which can limit the relevance of robustness evaluations, particularly when newer and more representative datasets are available.

Finally, the number of datasets used in each study also affects the credibility of robustness assessments. Among the 53 reviewed studies, 12 rely on a single dataset, 22 use two, and the remainder evaluate their methods on between three and six datasets (ref. Figure \ref{fig:eval_datasets}). The use of a single dataset may be justified in cases with a narrow application focus, (e.g., fraud detection or malware classification), where the goal is to assess feasibility in a domain-specific context rather than to establish broad robustness. Or in cases where the authors have to create their own dataset which is costly, such as the case in \cite{abusnaina2019examining}. However, it can also reduce the scope of the evaluation and increase the risk of overfitting to dataset-specific patterns. In contrast, studies that incorporate multiple datasets are better positioned to evaluate general robustness across a broader range of conditions. 

\clearpage
\begin{tcolorbox}[
colframe=black,          % Dark border for contrast
colback=gray!5,         % Light gray background (soft & elegant)
coltitle=white,         % White title text
colbacktitle=purple!50!black, % Deep purple title background
title=\bfseries  \Large Practical considerations, center title,% Bold title text
sharp corners=south,    % Sharp bottom corners
fonttitle=\bfseries     % Bold title
]

Across the surveyed literature, there is no consensus on the relative importance of the eight identified practical considerations. All studies address efficacy, with moderate attention given to feasibility and dataset suitability. In contrast, considerations such as defense awareness, transferability, plausibility, and semantic preservation are significantly less explored (each addressed in fewer than 19 of the 53 studies). \textit{Efficacy} is typically evaluated through classification error rates, with limited linkage to real-world objectives. There is no standardized metric for \textit{efficiency}, though runtime is the most commonly reported. Unlike in computer vision, \textit{transferability} is often treated as an auxiliary or secondary concern. \textit{Feasibility} constraints range from generic tabular rules (e.g., value ranges) to domain-specific and highly restrictive formulations. Addressing feasibility often intersects with research areas such as combinatorial optimization, constraint satisfaction, and neuro-symbolic learning. \textit{Semantic preservation} is rarely formalized as an optimization objective, and when it is, it is typically not evaluated post-attack. There is ongoing debate regarding its necessity for tabular data. A similar lack of consensus exists for \textit{plausibility}, where it remains unclear what constitutes a plausible adversarial example in tabular domains.  \textit{Defense strategies} are dominated by adversarial training, though tabular-specific techniques such as constraint-based augmentations are emerging. In terms of \textit{dataset suitability}, many studies rely on outdated or threat-irrelevant datasets, and often use only a single dataset. No standardized benchmarks exist, with 34 out of 61 datasets being used in just one study, a clear contrast to image domains where datasets like ImageNet serve as common baselines.
\end{tcolorbox}

\section{Discussion and Future directions}
\label{sec:discussions}

This review highlights that adversarial attacks for tabular ML remain a fragmented and relatively underdeveloped area. Despite a steady increase in publications, there is a lack of coordination across the research community. The diversity of publication venues and the absence of common benchmarks have contributed to limited interaction between studies. This is especially evident in learning-based approaches, where many papers rely on similar GAN-based architectures and constraints, yet rarely compare against each other or acknowledge existing work. As a result, similar contributions are often positioned as novel, even when the core techniques or assumptions are overlapping. This fragmentation is further reflected in the limited adoption of open-source practices: 21 out of 61 attacks do not release their code, making it difficult for subsequent work to build upon existing methods.

Another important challenge in this field is  is the lack of agreement on which practical aspects should guide evaluation. While efficacy is universally assessed, other factors—such as transferability, semantic preservation, plausibility, and defense awareness—receive inconsistent or minimal attention. Unlike computer vision, where evaluation frameworks are more standardized, the tabular domain lacks widely accepted guidelines. This raises the question of whether the community should establish shared evaluation criteria to better assess real-world applicability and practical impact. A key reason for this inconsistency is the conceptual ambiguity surrounding core notions like feasibility, semantics, and plausibility. Feasibility, for instance, is often prioritized, but its definition varies significantly—ranging from simple structural constraints to more complex domain-specific rules. Semantic preservation and plausibility are even less clearly defined, often only implied rather than formalized. Tools like Lp norms are occasionally used, but these are rarely grounded in real-world interpretability. To move forward, future research should aim to clarify the relationships among these concepts. For example, it remains unclear whether feasibility is a subset of plausibility, whether semantic preservation implies plausibility, or whether these are independent but overlapping goals. Greater conceptual clarity could guide both the formulation of optimization objectives and the development of more consistent, meaningful evaluation strategies.

Dataset usage also poses a significant barrier to progress. Many studies rely on outdated, narrowly focused, or under-documented datasets. Moreover, most experiments are conducted on a single dataset, which limits generalizability. Without common benchmarks, meaningful comparisons across studies are difficult. Recent initiatives such as TabularBench \cite{simonetto2024tabularbench} mark a step in the right direction, but their scope remains limited. Expanding the range of datasets, especially those from  high-stakes domains with clear adversarial risks and well-defined constraints, will be essential for improving the rigor and relevance of adversarial evaluations in tabular ML.

Another gap is the limited theoretical understanding of adversarial robustness in tabular settings. The literature is largely empirical, with few studies examining why certain attacks succeed or why specific models are more robust. By contrast, research in computer vision has proposed multiple theories of adversarial vulnerability including geometric, boundary-based, and lipshitz-based theories.
There is a clear opportunity to investigate similar questions in the context of tabular data, where the structure and data types differ significantly from images or text. Developing theoretical models that account for discrete variables, mixed feature types, and logic constraints could deepen our understanding and guide more effective defenses.

Current research on adversarial robustness in tabular data largely focuses on traditional supervised learning, especially classification, with models trained from scratch on single tasks. This overlooks emerging paradigms like pretraining, self-supervised learning, and multitask learning, which may introduce biases that influence robustness in ways not yet understood. Similarly, architectural choices such as model type or embedding strategies are rarely studied, despite their relevance. In parallel, modern techniques like context learning and retrieval-augmented inference, increasingly used in real-world systems, are beginning to appear in tabular settings, yet their adversarial properties remain completely unexplored. Beyond these learning paradigms and architectural shifts, tabular models are also being extended to more complex reasoning tasks such as question answering, report generation, and decision support. These emerging use cases introduce new types of vulnerabilities and evaluation challenges that current research has yet to address. Investigating how these trends interact with adversarial robustness is a key opportunity for future work.

\section{Conclusion}

Adversarial attacks on tabular ML models pose a significant yet understudied threat. While adversarial machine learning has matured in domains like vision and language, the tabular landscape remains fragmented, lacking shared methodologies, benchmarks, and foundational theory. This systematic literature review addresses that gap by analyzing 53 studies spanning diverse application areas. We identify key trends in publication activity, attack strategies, and the treatment of eight practical concerns: efficacy, efficiency, transferability, feasibility, plausibility, semantic preservation, defense awareness, and dataset suitability. Our findings show that while efficacy is consistently prioritized, other factors remain unevenly explored, with many conceptual foundations still unsettled. We also highlight emerging challenges related to model architectures, evolving task types, and new learning paradigms. As this field matures, progress will require both technical innovation and structural improvements. Establishing shared benchmarks, clarifying foundational concepts, and expanding the scope of threat models and learning settings will be essential for building a more coherent, rigorous, and impactful research agenda in adversarial robustness for tabular machine learning.

% \section{Acknowledgments}
% \begin{verbatim}
%   \begin{acks}
%   ...
%   \end{acks}
% \end{verbatim}

\bibliographystyle{ACM-Reference-Format}
\bibliography{references}

\newpage
\appendix
\section{Research trends}
\label{app:rq1}

\begin{table}[h!]
\centering
\caption{Metadata extracted from 53 papers under review. Citations obtained on 10/3/2025 through Google Scholar.}
\label{tab:metadata}
\footnotesize
\begin{tabular}
{@{}p{0.5cm}|p{0.5cm}p{0.5cm}|p{1.8cm}p{1.5cm}p{5cm}|p{1.2cm}|p{1.7cm}@{}} 
\toprule
\multicolumn{1}{c|}{\textbf{Ref.}} 
& \multicolumn{2}{c|}{\textbf{Year}} 
& \multicolumn{3}{c|}{\textbf{Venue}} 
& \multicolumn{1}{c|}{\textbf{Citations}} 
& \multicolumn{1}{c}{\textbf{Domain}} \\

& \multicolumn{1}{c}{\textbf{Pre.}} 
& \multicolumn{1}{c|}{\textbf{Pub.}} 
& \multicolumn{1}{c}{\textbf{Name}} 
& \multicolumn{1}{c}{\textbf{Type}} 
& \multicolumn{1}{c|}{\textbf{Domain}} 
& 
& \\
 \midrule
      \cite{lin2022idsgan} & 2018 & 2022 & PAKDD & Conference & Big Data, Data Mining, \& Knowledge Discovery & 426 & Cybersecurity \\ 
        \cite{ballet2019imperceptible} & 2019 & 2019 & arXiv & Pre-Print & Others & 115 & General \\ 
        \cite{shirazi2019adversarial} & 2019 & 2019 & DBSec & Conference & Cybersecurity & 45 & Cybersecurity \\ 
      \cite{usama2019generative} & 2019 & 2019 & IWCMC & Conference & Computer Networks \& Communication & 198 & Cybersecurity \\ 
      \cite{abusnaina2019examining} & 2019 & 2019 & DSC & Conference & Cybersecurity & 25 & Cybersecurity \\ 
      \cite{wang2020attackability} & 2020 & 2020 & KDD & Conference & Big Data, Data Mining, \& Knowledge Discovery & 22 & General \\ 
        \cite{sheatsley2020adversarial} & 2020 & 2020 & arXiv & Pre-Print & Others & 27 & General \\ 
       \cite{tian2020exploring} & 2020 & 2020 & ICPADS & Conference & System Design \& Engineering & 7 & General \\ 
       \cite{teuffenbach2020subverting} & 2020 & 2020 & MAKE & Conference & AI \& ML & 20 & Cybersecurity \\ 
        \cite{ghamizi2020search} & 2020 & 2020 & ESEC/FSE & Conference & Others & 31 & Finance \\ 
        \cite{alhajjar2021adversarial} & 2021 & 2020 & ESWA & Journal & System Design \& Engineering & 189 & Cybersecurity \\ 
        \cite{erdemir2021adversarial} & 2021 & 2021 & NeurIPS & Conference & AI \& ML & 34 & General \\ 
        \cite{gupta_adversarial_2021} & 2021 & 2021 & AI Safety & Workshop & AI \& ML & 30 & General \\ 
        \cite{gressel2021feature} & 2021 & 2021 & arXiv & Pre-Print & Others & 17 & General \\ 
        \cite{agarwal2021black} & 2021 & 2021 & MUFin21 & Workshop & Big Data, Data Mining, \& Knowledge Discovery & 13 & Finance \\ 
        \cite{cartella_adversarial_2021} & 2021 & 2021 & SafeAI & Workshop & AI \& ML & 95 & Finance \\ 
        \cite{kumar_evolutionary_2021} & 2021 & 2021 & ICMLA & Conference & AI \& ML & 17 & Finance \\ 
        \cite{shirazi2021directed} & 2021 & 2021 & JCS & Journal & Cybersecurity & 9 & Cybersecurity \\ 
       \cite{li2021conaml} & 2020 & 2021 & AsiaCCS & Conference & Cybersecurity & 76 & Other \\ 
        \cite{sheatsley2021robustness} & 2021 & 2021 & CCS & Conference & Cybersecurity & 35 & General \\ 
        \cite{zhao2021attackgan} & 2021 & 2021 & IIKI & Conference & Others & 50 & Cybersecurity \\ 
        \cite{de2021fixed} & 2021 & 2021 & ICML-ML4Cyber & Workshop & AI \& ML & 5 & Cybersecurity \\ 
       \cite{ju_robust_2022} & 2022 & 2022 & BigDIA & Conference & Big Data, Data Mining, \& Knowledge Discovery & 5 & General \\ 
        \cite{vitorino_adaptative_2022} & 2022 & 2022 & Future Internet & Journal & Computer Networks \& Communication & 33 & Cybersecurity \\ 
        \cite{mathov_not_2022} & 2020 & 2022 & KBS & Journal & Big Data, Data Mining, \& Knowledge Discovery & 32 & General \\ 
        \cite{nobi_adversarial_2022} & 2022 & 2022 & ITADATA & Conference & Big Data, Data Mining, \& Knowledge Discovery & 3 & Other \\ 
       \cite{chernikova_fence_2022} & 2019 & 2022 & TOPS & Journal & Cybersecurity & 52 & General \\ 
        \cite{chen2020generating} & 2022 & 2022 & TDSC & Journal & System Design \& Engineering & 69 & Cybersecurity \\ 
        \cite{simonetto2021unified} & 2021 & 2022 & IJCAI & Conference & AI \& ML & 24 & General \\ 
       \cite{simonetto2024towards} & 2022 & 2023 & AAAI & Conference & AI \& ML & 4 & General \\ 
        \cite{parfenov_investigation_2023} & 2023 & 2023 & USBEREIT & Conference & System Design \& Engineering & 1 & Other \\ 
       \cite{pandey_improving_2023} & 2023 & 2023 & ICAIF & Conference & Others & 3 & Finance \\ 
     \cite{malik_tetraa_2023} & 2023 & 2023 & IJCNN & Conference & AI \& ML & 0 & General \\ 
        \cite{kireev_adversarial_2023} & 2022 & 2023 & NDSS & Conference & Cybersecurity & 21 & General \\ 
        \cite{karumanchi_minimum_2023} & 2023 & 2023 & arXiv & Pre-Print & Others & 0 & General \\ 
        \cite{kong2023adversarial} & 2023 & 2023 & SMC & Journal & System Design \& Engineering & 6 & General \\ 
        \cite{nandy_non-uniform_2023} & 2023 & 2023 & CIKM & Conference & Big Data, Data Mining, \& Knowledge Discovery & 2 & General \\ 
        \cite{sun2023gpmt} & 2023 & 2023 & C \& S & Journal & Cybersecurity & 5 & Cybersecurity \\ 
        \cite{debicha2023adv} & 2023 & 2023 & C \& S & Journal & Cybersecurity & 35 & Cybersecurity \\ 
       \cite{asimopoulos_breaching_2023} & 2023 & 2023 & ARES & Conference & Cybersecurity & 9 & Cybersecurity \\ 
        \cite{duy2023investigating} & 2023 & 2023 & JISA & Journal & Cybersecurity & 17 & Cybersecurity \\ 
        \cite{xu2023probabilistic} & 2022 & 2023 & ICML & Conference & AI \& ML & 14 & General \\ 
        \cite{simonetto2024constrained} & 2023 & 2024 & NeurIPS & Conference & AI \& ML & 1 & General \\ 
       \cite{ben-tov_cafa_2024} & 2024 & 2024 & SP & Conference & Cybersecurity & 0 & General \\ 
    \cite{duan_attacking_2024} & 2024 & 2024 & TKDD & Journal & Big Data, Data Mining, \& Knowledge Discovery & 4 & Other \\ 
        \cite{dyrmishi2024deep} & 2024 & 2024 & ICMLC & Conference & AI \& ML & 1 & General \\ 
        \cite{simonetto2024towards} & 2024 & 2024 & NextGenAISafety & Workshop & AI \& ML & 1 & General \\ 
       \cite{concone2024adverspam} & 2024 & 2024 & TOPS & Journal & Cybersecurity & 7 & Cybersecurity \\ 
        \cite{alhussien2024constraining} & 2024 & 2024 & TNSM & Journal & Computer Networks \& Communication & 4 & Cybersecurity \\ 
        \cite{wang2024ids} & 2024 & 2024 & CVIDL & Conference & AI \& ML & 1 & Cybersecurity \\ 
  \cite{yuan2024multi} & 2022 & 2024 & DTRAP & Journal & Cybersecurity & 7 & Cybersecurity \\ 
        \cite{bai2024adversarial} & 2024 & 2024 & IPSA & Conference & Others & 0 & Other \\ 
     \cite{khazanchi2024mislead} & 2024 & 2024 & arXiv & Pre-Print & Others & 0 & General \\ 
     \bottomrule
    \end{tabular}
\end{table}

\newpage

\section{Practical considerations}
\label{app:rq3}

\begin{table}[h!]
\centering
\caption{Practical dimensions for evaluating adversarial attacks.}
\label{tab:rq3_marking_explained}
\footnotesize
\begin{tabular}{@{}p{2.5cm}p{3.7cm}p{3.7cm}p{3.7cm}@{}}
\toprule
\textbf{Dimension} & \textbf{Considered} & \textbf{Acknowledged} & \textbf{Not Considered} \\
\midrule
\textit{Efficacy} & 
Attack success is explicitly measured (e.g., evasion rate). & 
Explicitly noted as important but not addressed. & 
Not mentioned or evaluated. \\

\textit{Efficiency} & 
Efficiency (e.g., runtime, query count) is measured or optimized. & 
Explicitly noted as important but not addressed. & 
Not mentioned or evaluated. \\

\textit{Transferability} & 
Attack is tested across models to assess generalization. & 
Explicitly noted as important but not addressed. & 
Not mentioned or evaluated. \\

\textit{Feasibility} & 
Realistic constraints enforced (e.g., clipping, repair, functional checks). & 
Explicitly noted as important but not addressed. & 
Not mentioned or evaluated. \\

\textit{Semantic Preservation} & 
Actively discussed and validated. & 
Explicitly noted as important but not addressed. This includes cases where the authors do not agree with the treatment of this aspect in the current literature or its application to tabular data. & 
Not discussed. Includes methods using $L_p norms$ or GANs without semantic checks or definition of semantic preservation. \\

\textit{Plausibility} & 
Actively discussed and evaluated (i.e automated checks or human review). & 
Explicitly noted as important, but not addressed. This includes cases where the authors do not agree with the treatment of this aspect in the current literature or its application to tabular data.& 
Not discussed. Includes use of $L_norms$ or GANs without clarifying relation to plausibility. \\

\textit{Defense Awareness} & 
Evaluated against one or more defense mechanisms. & 
Defenses mentioned in limitations or future work. & 
Not mentioned or evaluated. \\

\textit{Dataset Suitability} & 
Uses well-justified or multiple datasets aligned with the threat model. & 
Plans to improve dataset selection mentioned. & 
Not discussed. Includes use of a single or irrelevant dataset without threat model justification. \\
\bottomrule
\end{tabular}
\end{table}

\begin{table}[h!]
\small
\caption{Practical considerations of adversarial attacks on tabular ML. 
\checkmark: Considered, $\triangle$: Acknowledged, $\times$: Not considered.}
\label{tab:rq3_detailed}
\resizebox{0.85\textwidth}{!}{
\begin{tabular}{*{9}{c}} 
\toprule
\textbf{Ref.} & \textbf{Efficacy} & \textbf{Efficiency} & \textbf{Transferability} & \textbf{Feasibility} & \textbf{Semantics} & \textbf{Plausibility} & \textbf{Defense} & \textbf{Dataset} \\
\midrule
\cite{lin2022idsgan}                                                 & \checkmark & $\times$    & $\times$         & \checkmark  & $\times$    & $\times$      & $\times$    & $\times$    \\
 \cite{ballet2019imperceptible}                                                     & \checkmark & $\triangle$  & $\triangle$       & \checkmark  & $\times$    & \checkmark   & $\times$    & \checkmark \\
\cite{shirazi2019adversarial}                                                                                 & \checkmark & $\times$    & $\times$         & \checkmark  & $\times$    & \checkmark   & $\triangle$  & \checkmark \\
 \cite{usama2019generative}                     & \checkmark & $\times$    & $\times$         & \checkmark  & $\times$    & $\times$      & \checkmark & $\times$    \\
 \cite{abusnaina2019examining}                                                   & \checkmark & $\times$    & $\times$         & \checkmark  & $\times$    & $\times$      & \checkmark & $\times$    \\
\cite{sheatsley2020adversarial}                                                                                               & \checkmark & $\times$    & \checkmark      & \checkmark  & $\times$    & $\triangle$    & $\triangle$  & \checkmark \\
 \cite{tian2020exploring}                                     & \checkmark & $\triangle$  & \checkmark      & \checkmark  & $\times$    & \checkmark   & $\triangle$  & $\times$    \\
 \cite{teuffenbach2020subverting}                          & \checkmark & $\times$    & $\triangle$       & \checkmark  & \checkmark & $\triangle$    & $\times$    & \checkmark \\
\cite{ghamizi2020search}                                               & \checkmark & $\triangle$  & $\triangle$       & \checkmark  & $\times$    & $\times$      & \checkmark & \checkmark \\
\cite{alhajjar2021adversarial}                                                                     & \checkmark & $\times$    & \checkmark      & \checkmark  & $\times$    & $\times$      & $\times$    & \checkmark \\
\cite{chen2020generating}                          & \checkmark & $\triangle$  & \checkmark      & \checkmark  & \checkmark & $\times$      & \checkmark & \checkmark \\
\cite{wang2020attackability}                                                            & \checkmark & \checkmark & $\times$         & $\triangle$   & $\times$    & $\times$      & $\times$    & $\times$    \\
 \cite{erdemir2021adversarial}                                                                                   & \checkmark & $\times$    & $\times$         & \checkmark  & $\times$    & $\times$      & $\times$    & \checkmark \\
\cite{gupta_adversarial_2021}                                                                       & \checkmark & $\times$    & $\times$         & $\times$     & $\times$    & $\times$      & $\times$    & \checkmark \\
 \cite{gressel2021feature}                                                                    & \checkmark & $\triangle$  & $\times$         & \checkmark  & $\times$    & $\triangle$    & $\times$    & $\times$    \\
\cite{agarwal2021black}                                                               & \checkmark & $\times$    & $\times$         & \checkmark  & $\times$    & $\times$      & $\times$    & \checkmark \\
 \cite{cartella_adversarial_2021}                                                 & \checkmark & $\times$    & \checkmark      & \checkmark  & $\times$    & \checkmark   & $\triangle$  & $\triangle$  \\
 \cite{kumar_evolutionary_2021}                                                                                     & \checkmark & \checkmark & $\times$         & \checkmark  & $\times$    & $\times$      & $\times$    & \checkmark \\
\cite{shirazi2021directed}                                                                                & \checkmark & $\times$    & \checkmark      & \checkmark  & $\times$    & \checkmark   & $\triangle$  & \checkmark \\
\cite{li2021conaml}                                                             & \checkmark & \checkmark & $\triangle$       & \checkmark  & $\times$    & $\times$      & $\triangle$  & \checkmark \\
 \cite{sheatsley2021robustness}                                                                                                 & \checkmark & \checkmark & $\times$         & \checkmark  & \checkmark & $\times$      & $\times$    & \checkmark \\
\cite{zhao2021attackgan}                                                  & \checkmark & $\times$    & $\times$         & \checkmark  & $\times$    & $\times$      & $\triangle$  & $\times$    \\
 \cite{de2021fixed}                                                    & \checkmark & \checkmark & $\times$         & $\times$     & $\times$    & $\times$      & \checkmark & $\times$    \\
 \cite{ju_robust_2022}                                             & \checkmark & $\times$    & \checkmark      & $\times$     & $\times$    & \checkmark   & $\times$    & $\times$    \\
\cite{vitorino_adaptative_2022}                                          & \checkmark & \checkmark & $\times$         & \checkmark  & $\times$    & $\times$      & \checkmark & \checkmark \\
\cite{mathov_not_2022}                                                  & \checkmark & $\times$    & \checkmark      & \checkmark  & \checkmark & $\times$      & $\times$    & $\times$    \\
\cite{nobi_adversarial_2022}                                                                              & \checkmark & $\times$    & $\times$         & \checkmark  & $\times$    & $\times$      & $\triangle$  & \checkmark \\
 \cite{chernikova_fence_2022}                                                           & \checkmark & $\times$    & \checkmark      & \checkmark  & $\times$    & $\times$      & \checkmark & \checkmark \\
\cite{simonetto2021unified}                                                      & \checkmark & $\times$    & $\times$         & \checkmark  & $\times$    & $\times$      & \checkmark & \checkmark \\
\cite{simonetto2024constrained}    & \checkmark & \checkmark & \checkmark      & \checkmark  & $\times$    & $\times$      & \checkmark & $\times$    \\
 \cite{bao2023towards}                                             & \checkmark & \checkmark & $\times$         & $\times$     & $\times$    & $\times$      & $\triangle$  & $\times$    \\
 \cite{parfenov_investigation_2023}                                                       & \checkmark & $\times$    & $\times$         & \checkmark  & $\times$    & $\times$      & $\times$    & $\times$    \\
\cite{pandey_improving_2023}                               & \checkmark & \checkmark & $\times$         & $\times$     & $\times$    & \checkmark   & \checkmark & \checkmark \\
 \cite{malik_tetraa_2023}                                                         & \checkmark & \checkmark & $\times$         & \checkmark  & $\times$    & \checkmark   & $\times$    & \checkmark \\
\cite{kireev_adversarial_2023}                                                                & \checkmark & \checkmark & $\times$         & \checkmark  & $\triangle$  & $\triangle$    & \checkmark & \checkmark \\
 \cite{karumanchi_minimum_2023}                                                                                       & \checkmark & \checkmark & $\times$         & \checkmark  & $\times$    & $\times$      & $\times$    & \checkmark \\
\cite{kong2023adversarial}                                                                   & \checkmark & $\times$    & $\times$         & $\times$     & $\times$    & $\times$      & $\triangle$  & \checkmark \\
 \cite{nandy_non-uniform_2023}                                                                      & \checkmark & $\triangle$  & $\times$         & \checkmark  & $\times$    & $\times$      & \checkmark & $\times$    \\
\cite{sun2023gpmt}                                 & \checkmark & $\times$    & $\times$         & \checkmark  & \checkmark & $\times$      & \checkmark & \checkmark \\
\cite{debicha2023adv}                                                 & \checkmark & \checkmark & \checkmark      & \checkmark  & \checkmark & $\times$      & \checkmark & \checkmark \\
 \cite{duy2023investigating} & \checkmark & $\triangle$  & $\triangle$       & \checkmark  & $\times$    & \checkmark   & \checkmark & \checkmark \\
 \cite{xu2023probabilistic}                                                                       & \checkmark & \checkmark & $\times$         & $\times$     & $\times$    & $\times$      & \checkmark & \checkmark \\
 \cite{ben-tov_cafa_2024}                                        & \checkmark & \checkmark & $\times$         & \checkmark  & $\times$    & $\times$      & $\triangle$  & \checkmark \\
 \cite{duan_attacking_2024}                                                         & \checkmark & \checkmark & \checkmark      & \checkmark  & $\times$    & \checkmark   & \checkmark & \checkmark \\
 \cite{dyrmishi2024deep}                                                      & \checkmark & \checkmark & $\times$         & \checkmark  & $\times$    & $\times$      & $\times$    & $\times$    \\
 \cite{simonetto2024towards}                                                                       & \checkmark & $\triangle$  & $\times$         & \checkmark  & $\times$    & $\times$      & $\times$    & $\times$    \\
\cite{concone2024adverspam}                                                                & \checkmark & $\triangle$  & \checkmark      & \checkmark  & $\times$    & \checkmark   & \checkmark & $\triangle$  \\
\cite{papernot2016transferability}                            & \checkmark & \checkmark & \checkmark      & \checkmark  & $\times$    & $\times$      & \checkmark & $\times$    \\
\cite{wang2024ids}                                     & \checkmark & $\times$    & $\times$         & \checkmark  & $\times$    & $\times$      & $\times$    & $\times$    \\
 \cite{yuan2024multi}        & \checkmark & $\triangle$  & $\times$         & \checkmark  & $\times$    & $\triangle$    & $\triangle$  & \checkmark \\
\cite{khazanchi2024mislead}                                     & \checkmark & $\triangle$  & $\times$         & \checkmark  & $\times$    & $\times$      & $\triangle$  & $\times$    \\
\cite{asimopoulos_breaching_2023}       & \checkmark & $\times$    & $\times$         & $\times$     & $\times$    & $\times$      & $\triangle$  & $\times$    \\
 \cite{bai2024adversarial}                                       & \checkmark & $\times$    & \checkmark      & \checkmark  & $\times$    & $\times$      & $\triangle$  & $\times$    \\
\bottomrule
\end{tabular}
}
\end{table}

\newpage

\begin{table}[h!]
\centering
\footnotesize
\caption{Overview of datasets used for evaluation along with relevant statistics. Here, \#F denotes the total number of features, \#C the number of categorical features, \#N the number of numerical features, and \#Usages the number of papers in our survey that use each dataset as part of their evaluation methodology.}
\label{tab:datasets}
\begin{tabular}{p{2.5cm} p{2.5cm} p{1.5cm} p{0.5cm} p{0.5cm} p{0.5cm} p{0.7cm}p{3.3cm}}
\toprule
\textbf{Dataset} & \textbf{Short description} & \textbf{Domain} & \textbf{\#F} & \textbf{\#C} & \textbf{\#N} & \textbf{\#Usages } & \textbf{Used by} \\
\midrule
KDD99 \cite{kdd_cup_1999_data_130} & Botnet traffic & Cybersecurity & 41 & 34 & 7 & 1 & \cite{usama2019generative} \\
NSL-KDD \cite{tavallaee2009detailed} & Improved KDD99 & Cybersecurity & 41 & 34 & 7 & 8 & \cite{lin2022idsgan, sheatsley2020adversarial, tian2020exploring, teuffenbach2020subverting, alhajjar2021adversarial, sheatsley2021robustness, zhao2021attackgan, wang2024ids} \\
CTU-13 \cite{garcia2014empirical} & Botnet traffic & Cybersecurity & 13 & 2 & 11 & 6 & \cite{chernikova_fence_2022, sun2023gpmt, debicha2023adv, simonetto2021unified, simonetto2024constrained, simonetto2024towards} \\
ISCX-Botnet \cite{beigi2014towards} & Botnet traffic & Cybersecurity & 33 & 6 & 27 & 1 & \cite{sun2023gpmt} \\
UNSW-NB15 \cite{moustafa2015unsw} & Malicious traffic & Cybersecurity & 48 & 2 & 46 & 3 & \cite{sheatsley2020adversarial, alhajjar2021adversarial, alhussien2024constraining} \\
DDoS-A \cite{niyaz2016deep} & DDoS traffic & Cybersecurity & 54 & 0 & 54 & 1 & \cite{abusnaina2019examining} \\
CIC-IDS2017 \cite{sharafaldin2018toward} & Malicious traffic & Cybersecurity & 78 & 0 & 78 & 2 & \cite{teuffenbach2020subverting, vitorino_adaptative_2022} \\
CIC-IDS2018 \cite{sharafaldin2018toward} & Malicious traffic & Cybersecurity & 77 & 1 & 76 & 2 & \cite{debicha2023adv, duy2023investigating} \\
IoT-23 \cite{garcia2020iot} & Malicious IoT traffic & Cybersecurity & 20 & 4 & 16 & 2 & \cite{vitorino_adaptative_2022, alhussien2024constraining} \\
InSDN \cite{elsayed2020insdn} & Malicious SDN traffic & Cybersecurity & 56 & 1 & 55 & 1 & \cite{duy2023investigating} \\
IPS-A \cite{wang2020attackability} & Intrusion Prevention & Cybersecurity & 1103 & 1103 & 0 & 2 & \cite{xu2023probabilistic, wang2020attackability} \\
IEC 60870-5-104 \cite{radoglou2021modeling} & Malicious IoT traffic & Cybersecurity & -- & -- & -- & 1 & \cite{asimopoulos_breaching_2023} \\
Phishing-A \cite{mohammad2012assessment} & Website phishing & Cybersecurity & 17 & 17 & -- & 2 & \cite{shirazi2019adversarial, shirazi2021directed} \\
Phishing-B \cite{abdelhamid2014phishing} & Website phishing & Cybersecurity & 16 & 16 & -- & 2 & \cite{shirazi2019adversarial, shirazi2021directed} \\
ISCX-URL2016 \cite{mamun2016detecting} & Malicious URLs & Cybersecurity & 79 & -- & -- & 1 & \cite{gressel2021feature} \\
DeltaPhish \cite{corona2017deltaphish} & Website phishing & Cybersecurity & 11 & 2 & 9 & 1 & \cite{yuan2024multi} \\
Phishing-C \cite{shirazi2018kn0w} & Website phishing & Cybersecurity & 8 & 4 & 4 & 2 & \cite{shirazi2019adversarial, shirazi2021directed} \\
Phishing-D \cite{chiew2019new} & Website phishing & Cybersecurity & 48 & 29 & 19 & 3 & \cite{sheatsley2021robustness, ben-tov_cafa_2024, gressel2021feature} \\
Phishing-E \cite{lee2020building} & Website phishing & Cybersecurity & 51 & -- & -- & 1 & \cite{gressel2021feature} \\
Phishing-F \cite{hannousse2021towards} & Website phishing & Cybersecurity & 87 & 26 & 61 & 4 & \cite{simonetto2021unified, simonetto2024constrained, dyrmishi2024deep, simonetto2024towards} \\
Phishing-G \cite{gressel2021feature} & Website phishing & Cybersecurity & 52 & 3 & 49 & 2 & \cite{karumanchi_minimum_2023, gressel2021feature} \\
Phishing-H \cite{van2021combining} & Website phishing & Cybersecurity & -- & -- & -- & 1 & \cite{yuan2024multi} \\
Spam-base \cite{spambase_94} & Email spam & Cybersecurity & 57 & 0 & 57 & 1 & \cite{ju_robust_2022} \\
Twitter-A \cite{lee2011seven} & Twitter spam & Cybersecurity & 16 & 0 & 16 & 1 & \cite{gupta_adversarial_2021} \\
Twitter-B \cite{gilani2017classification} & Twitter spam & Cybersecurity & 15 & 1 & 14 & 1 & \cite{kireev_adversarial_2023} \\
Twitter-C \cite{fazil2018hybrid} & Twitter spam & Cybersecurity & 17 & 0 & 17 & 1 & \cite{concone2024adverspam} \\
EMBER \cite{anderson2018ember} & Windows malware & Cybersecurity & 2381 & 0 & 2381 & 1 & \cite{erdemir2021adversarial} \\
DroidFusion \cite{yerima2018droidfusion} & Android malware & Cybersecurity & 350 & 350 & 0 & 1 & \cite{karumanchi_minimum_2023} \\
Mal-A \cite{dyrmishi2023empirical} & Windows malware & Cybersecurity & 24222 & -- & -- & 1 & \cite{simonetto2021unified} \\
Australian Credit \cite{statlog_australian_credit_approval_143} & Credit risk & Finance & 14 & 8 & 6 & 2 & \cite{ballet2019imperceptible, agarwal2021black} \\
German Credit \cite{statlog_german_credit_data_144} & Credit risk & Finance & 20 & 13 & 7 & 3 & \cite{ballet2019imperceptible, gupta_adversarial_2021, cartella_adversarial_2021} \\
Credit Card \cite{yeh2009comparisons} & Credit default & Finance & 23 & 4 & 19 & 3 & \cite{ballet2019imperceptible, agarwal2021black, malik_tetraa_2023} \\
Bank Marketing \cite{moro2014data} & Direct marketing & Finance & 16 & 8 & 8 & 2 & \cite{ben-tov_cafa_2024, khazanchi2024mislead} \\
Loan-ICL \cite{loan-default-prediction} & Credit risk & Finance & 778 & -- & 1 & 1 & \cite{kong2023adversarial} \\
ECD \cite{dal2015adaptive} & Credit card fraud & Finance & 30 & 0 & 30 & 3 & \cite{kumar_evolutionary_2021, pandey_improving_2023, malik_tetraa_2023} \\
HomeCredit \cite{home-credit-default-risk} & Credit scoring & Finance & 974 & -- & -- & 2 & \cite{kireev_adversarial_2023, mathov_not_2022} \\
Heloc \cite{fico_heloc} & Credit risk & Finance & 24 & 8 & 16 & 1 & \cite{dyrmishi2024deep} \\
IEEE-CIS Fraud \cite{ieee-fraud-detection} & Credit card fraud & Finance & 433 & 48 & -- & 4 & \cite{cartella_adversarial_2021, kumar_evolutionary_2021, pandey_improving_2023, kireev_adversarial_2023} \\
LCLD \cite{lending_club_dataset} & Peer-to-peer loans & Finance & 151 & -- & -- & 5 & \cite{ballet2019imperceptible, ghamizi2020search, simonetto2021unified, simonetto2024constrained, simonetto2024towards} \\
Breast Cancer \cite{breast_cancer_14} & Cancer statistics & Healthcare & 9 & 9 & 0 & 1 & \cite{ju_robust_2022} \\
EHR \cite{ma2018risk} & Electronic health records & Healthcare & 4130 & 4130 & 0 & 2 & \cite{bao2023towards, wang2020attackability} \\
WiDS \cite{lee2020wids} & Patients ICU data & Healthcare & 109 & 9 & 100 & 4 & \cite{simonetto2024constrained, dyrmishi2024deep, simonetto2024towards, mathov_not_2022} \\
iPinYou \cite{liao2014ipinyou} & Click-through rate & Other & 21 & 2 & 19 & 1 & \cite{duan_attacking_2024} \\
Criteo \cite{criteo-display-ad-challenge} & Click-through rate & Other & 39 & 26 & 13 & 2 & \cite{nandy_non-uniform_2023, duan_attacking_2024} \\
Avazu \cite{avazu-ctr-prediction} & Click-through rate & Other & 21 & 8 & 13 & 1 & \cite{duan_attacking_2024} \\
Iris \cite{fisher1936use} & Flower characteristics & Other & 4 & 0 & 4 & 2 & \cite{khazanchi2024mislead, ju_robust_2022} \\
Wine \cite{wine_109} & Chemical characteristics & Other & 13 & 0 & 13 & 1 & \cite{ju_robust_2022} \\
Adult \cite{adult_2} & Census data & Other & 14 & 8 & 6 & 2 & \cite{ben-tov_cafa_2024, gressel2021feature} \\
Magic Telescope \cite{magic_gamma_telescope_159} & Particle physics & Other & 10 & 0 & 10 & 1 & \cite{ju_robust_2022} \\
NASA C-MAPSS \cite{saxena2008damage} & Engine Degradation & Other & 26 & 0 & 26 & 1 & \cite{kong2023adversarial} \\
Rob-Navigation \cite{wall-following_robot_navigation_data_194} & Robot movement & Other & 24 & 0 & 24 & 1 & \cite{ju_robust_2022} \\
MovieLens \cite{harper2015movielens} & Movie recommendations & Other & 3 & 0 & 3 & 1 & \cite{nandy_non-uniform_2023} \\
GAS-IDS \cite{morris2015industrial} & Gas pipeline network & Other & 15 & 5 & 10 & 1 & \cite{bai2024adversarial} \\
FSP \cite{uciml_faulty_steel_plates} & Faulty steel plates & Other & 27 & 0 & 27 & 1 & \cite{dyrmishi2024deep} \\
SWaT \cite{goh2017dataset} & Secure Water Treatment & Other & 51 & 26 & 25 & 1 & \cite{li2021conaml} \\
U.S. Lodging \cite{airbnb_dataset} & AirBnB data & Other & 25 & 8 & 17 & 1 & \cite{mathov_not_2022} \\
DeepMimo \cite{alkhateeb2019deepmimo} & Wireless communication & Other & 5 & 0 & 5 & 1 & \cite{parfenov_investigation_2023} \\
ICS-A \cite{chen2020generating} & Water tank comm. & Other & 12 & 0 & 12 & 1 & \cite{chen2020generating} \\
IEEE-10-39 \cite{li2021conaml} & Power flow data & Other & 46 & 0 & 46 & 1 & \cite{li2021conaml} \\
u5k-r5k-auth12k \cite{nobi2022toward} & User access control & Other & 16 & 8 & 8 & 1 & \cite{nobi_adversarial_2022} \\
Ammonia \cite{kong2023adversarial} & Ammonia $CO_2$ synthesis & Other & 11 & 0 & 11 & 1 & \cite{kong2023adversarial} \\
\bottomrule
\end{tabular}
\end{table}

\end{document}